\documentclass[journal]{IEEEtran}

\usepackage{graphics} 
\usepackage{epsfig} 
\usepackage{amsmath,amsfonts,amssymb,bm} 
\usepackage{subfigure,cases,multirow}
\usepackage{algorithm,algpseudocode,algorithmicx}
\usepackage[colorlinks,citecolor=blue,urlcolor=black,linkcolor=blue]{hyperref}
\usepackage{accents}
\usepackage[displaymath,mathlines,switch]{lineno}
\usepackage[mathscr]{eucal}
\usepackage{booktabs}
\usepackage{color}

\makeatletter
\def\wideubar{\underaccent{{\cc@style\underline{\mskip10mu}}}}
\def\ubar{\underaccent{{\cc@style\underline{\mskip6mu}}}}
\makeatother

\makeatletter
\def\widebar{\accentset{{\cc@style\underline{\mskip10mu}}}}
\makeatother

\ifCLASSOPTIONcompsoc
  \usepackage[nocompress]{cite}
\else
  \usepackage{cite}
\fi

\ifCLASSINFOpdf
\else
\fi
\def\bR{\mathbb R}
\def\bS{\mathbb S}

\def\bZ{\mathbb Z}

\def\rU{\Phi}

\def\cB{\mathcal B}
\def\cC{\mathcal C}
\def\cM{\mathcal M}

\def\cN{\mathcal N}
\def\cF{\mathcal F}
\def\cR{\mathcal R}

\def\cH{\mathcal H}
\def\cS{\mathcal S}

\def\cJ{\mathcal J}
\def\cW{\mathcal W}

\def\cT{\mathfrak M}

\def\rG{\mathscr G}
\def\rD{\mathscr D}
\def\rV{\mathscr V}
\def\rE{\mathscr E}
\def\rA{\mathscr A} 
\def\kD{\mathfrak D}
\def\kF{\mathfrak F}

\def\kN{\mathfrak N}

\def\kR{\mathfrak R}

\def\fw{\mathbf w}

\def\fp{\mathbf p}
\def\fu{\mathbf u}
\def\fv{\mathbf v}
\def\fI{\mathbf I}
\def\fn{\mathbf n}

\def\GT{\emph{GT}}

\DeclareMathOperator\Lip{Lip}
\DeclareMathOperator\sign{Sign}
\DeclareMathOperator\Id{I_d}
\DeclareMathOperator\Vor{Vor}

\def\<{\langle}
\def\>{\rangle}

\newtheorem{proposition}{Proposition}

\newlength\savewidth
\newcommand\shline{\noalign{\global\savewidth\arrayrulewidth\global\arrayrulewidth 1.0pt}\hline\noalign{\global\arrayrulewidth\savewidth}}

\newlength\savedwidth

\begin{document}
\title{A Generalized Asymmetric Dual-front Model for Active Contours and Image Segmentation}

\author{Da~Chen,~Jack~Spencer, Jean-Marie Mirebeau, Ke Chen, Minglei Shu and Laurent D. Cohen, ~\IEEEmembership{Fellow,~IEEE}
       
\thanks{Da~Chen and Minglei Shu are with Qilu University of Technology (Shandong Academy of Sciences),  Shandong Artificial Intelligence Institute, China (e-mails: dachen.cn@hotmail.com; shuml@sdas.org) (Minglei Shu is the corresponding author)}
\thanks{Jack Spencer is with the Translational Research Exchange @ Exeter, Living Systems Institute, University of Exeter, Exeter EX4 4QD, U.K. (e-mail: j.a.spencer@exeter.ac.uk).}
\thanks{Ke Chen is with Department of Mathematical Sciences, The University of Liverpool, UK. (e-mail: k.chen@liverpool.ac.uk)}
\thanks{Jean-Marie Mirebeau is with Laboratoire de math\'ematiques d'Orsay, CNRS,  Universit\'e Paris-Sud, Universit\'e Paris-Saclay, 91405 ORSAY, France.(e-mail: jean-marie.mirebeau@math.u-psud.fr)}
\thanks{Laurent D. Cohen is with University Paris Dauphine, PSL Research University, CNRS, UMR 7534, CEREMADE, 75016 Paris, France. (e-mail: cohen@ceremade.dauphine.fr)}
}
\markboth{Journal of \LaTeX\ Class Files,~Vol.~14, No.~8, August~2018}
{Shell \MakeLowercase{\textit{et al.}}: Bare Demo of IEEEtran.cls for Journals}
\maketitle

\begin{abstract}
The Voronoi diagram-based dual-front scheme is known as a powerful and efficient technique for addressing the image segmentation and domain partitioning problems. In the basic formulation of existing dual-front approaches, the evolving contour can be considered as the interfaces of adjacent Voronoi regions. Among these dual-front models, a crucial ingredient is regarded as the geodesic metrics by which the geodesic distances and the corresponding Voronoi diagram can be estimated. In this paper, we introduce a new dual-front model based on asymmetric quadratic metrics. These metrics considered are built by the integration of the image features and a vector field derived from the evolving contour.  The use of the asymmetry enhancement can reduce the risk for the segmentation contours being stuck at false positions,  especially when the initial curves are far away from the target boundaries or the images have complicated intensity distributions. Moreover, the proposed dual-front model can be applied for image segmentation in conjunction with various region-based homogeneity terms. The numerical experiments on both synthetic and real images show that the proposed dual-front model indeed achieves encouraging results. 
\end{abstract}

\begin{IEEEkeywords}
Eikonal equation, asymmetric quadratic metric, Voronoi diagram, active contours, image segmentation, fast marching m ethod.
\end{IEEEkeywords}
\IEEEpeerreviewmaketitle

\section{Introduction}
\label{sec:intro}
Active contour models have been dedicated to suitably address many image segmentation tasks in a wide variety of computer vision and image analysis scenarios in the past three decades. Basically, image segmentation tasks carried out by an active contour model is usually implemented via a curve  evolution scheme. In essence, this scheme can be governed by the contour representation methods in conjunction with the image data-based energy functionals. 

Since the original work of the snakes model~\cite{kass1988snakes}, great efforts have contributed to investigate suitable energy functionals to satisfy a wide variety of image segmentation situations. In other words, the associated active contour approaches attempt to find proper ways on how to utilize the image features to define the objective boundaries. The edge-based features such as the image gradients are  widely utilized by many active contour approaches. Interesting examples for edge-based active contour approaches may include the geometric  models~\cite{caselles1993geometric,caselles1997geodesic,malladi1995shape,yezzi1997geometric}, the external force-based models~\cite{cohen1991active,cohen1993finite,xu1998snakes,xie2008mac} and the models relying on minimal cost paths~\cite{cohen1997global,appleton2005globally,mishra2011decoupled}. In~\cite{kimmel2003regularized,melonakos2008finsler}, the  edge anisotropy features are taken into account , leading to more general geometric active contours models. The use of edge features removes the effects from image gray level or color homogeneities, yielding practical applications.  However, the contour evolution driven by edge-based features may be trapped into unexpected local minima due to the presence of spurious edges probably generated by noises.
 
The region-based active contour models usually derive the gradient flows by minimizing energy functionals involving region-based homogeneity terms. The Mumford-Shah functional~\cite{mumford1989optimal} invokes a piecewise smooth fitting function to approximate the image data. The approximation is carried via a region-based term that characterizes the errors between the image gray levels and the the data fitting function. Following the Mumford-Shah piecewise smooth functional, a series of region-based active contour models were introduced to address various image segmentation issues. These models either consider the suitable variants of the Mumford-Shah functional~\cite{zhu1996region,chan2001active,brox2009local,li2008minimization} or introduce practical avenues to search for the solutions~\cite{vese2002multiphase,tsai2001curve,bresson2007fast, chambolle2012convex,grady2009piecewise}. The histograms or probability density functions of the image features such as image colors, gray levels and gradients as reviewed in~\cite{cremers2007review},  are often used to build nonparametric energy functionals, which can avoid to assume prior distribution of image intensities as introduced in the literature~\cite{ni2009local,michailovich2007image}. Recently, a new type of region-based active contour models based on the pairwise similarity-based energy functionals were introduced in~\cite{jung2012nonlocal,sumengen2006graph}, which have obtained encouraging results.
The active contour models mentioned above are obviously not exhaustive, interesting and effective  approaches  may include~\cite{zhang2019resls,min2018late,sundaramoorthi2007sobolev,li2011level,chen2016finsler,chen2017global}.

\noindent\emph{Contour Representation}.
The representation for evolving contour is a fundamental and challenging problem in active contour approaches. The parametrized contour method has been used in many approaches~\cite{kass1988snakes,cohen1991active,xu1998snakes}, due to its low computation complexity.  However, this method often suffers from the self-crossing during the contour evolution. In order to obtain expected results,  additional procedures are often required to alleviate that issue~\cite{nakhmani2012self}.  

The level set scheme~\cite{osher1988fronts} has been broadly exploited to search for suitable solutions to active contour evolution~\cite{caselles1997geodesic,yezzi1997geometric,malladi1995shape} due to its solid mathematical background and the rich numerical implementation methods~\cite{zhao1996variational,li2010distance,estellers2012efficient,saye2012analysis}. By the level set framework, the contour evolution can be regarded as a way of updating a Lipchitz function $\phi:\Omega\to\bR$, where $\Omega\subset\bR^d$ is an open and bounded domain of dimension $d=2,3$. In its basic formulation, the boundary $\partial\cR$ of a region $\cR\subset\Omega$ can be implicitly represented as the zero-level set of $\phi$ such that $\partial\cR=\{x\in\Omega;~\phi(x)=0\}$.
In this case, a point $x$ is inside $\cR$ if $\phi(x)<0$ and outside $\cR$ for $\phi(x)>0$. 
The variational level set method~\cite{zhao1996variational,chan2001active} made use of the Heaviside function $H:\bR\to\{0,1\}$ in conjunction with $\phi$ to assign a label to each point $x$, where $H(\phi(x))=1$ implies that $x\in\cR$. Nevertheless, $H(\phi)$ acts as the characteristic function of the region $\cR$.  
 However, minimizing an energy functional with respect to a level set function (LSF) $\phi$ corresponds to a local minimum, thus increasing the risk of finding unexpected image segmentation. Moreover, a small time step is usually adopted in order to find stable numerical solutions to the level set evolution equations, which will increase computation cost.

The convex relaxation minimization framework~\cite{chan2006algorithms,bae2011global,chambolle2011first} was introduced to address the above issues of the level set method in some extent. The convex relaxation schemes are able to find the global minimum of an active contour energy. Specifically, the active contour energy functionals, which are usually comprised of a region-based term and a regularization term, are reformulated by replacing the binary-valued characteristic function $H(\phi)$ using a function $\varphi:\Omega\to[0,1]$. As a consequence, the segmented regions can be derived by thresholding the solutions $\varphi$. The convex relaxation framework has been proven to achieve lower computation complexity  than that of the level set scheme and has been successfully applied in many segmentation tasks~\cite{bresson2007fast,goldstein2010geometric}.

Image segmentation approaches based on the concept of Voronoi diagram have obtained promising segmentation results in various segmentation tasks. Among them, the Voronoi diagram can be constructed from several user-provided scribbles respectively placed in the foreground and background regions, as introduced in~\cite{arbelaez2004energy,bai2009geodesic,chen2018asymmetric}. Therefore,  the user intervention can be naturally incorporated into the segmentation procedure. 
Alternatively, the Voronoi diagram can be investigated for active contour evolution, for which the basic idea is to represent a contour segment as the interface of two adjacent Voronoi regions. Along this research line, Voronoi regions can be generated through the  offset lines of the evolving contour using either geodesic distances~\cite{li2007local} or with Euclidean distances~\cite{saye2012analysis,dubrovina2015multi}. Specifically, the Voronoi implicit interface (VII) scheme~\cite{saye2012analysis} is a variant of the original level set method~\cite{osher1988fronts},  where the LSF $\phi$ is set as a non-negative Euclidean distance map associated to the evolving contour. 
In the context of multiphase segmentation, unlike  the traditional level set method or the convex relaxation method, which require  multiple $\phi$ or $\varphi$ to characterize all regions, only one single LSF is sufficient for the VII scheme-based dual-front model~\cite{dubrovina2015multi} to represent all disjoint regions. However, the VII model still needs to address the classical level set evolution problem in order to evolve the offset lines of the current contour, which may suffer from the issues of, for examples,  high computation cost and sensitivity to the parameters. 
Li and Yezzi introduced a dual-front model~\cite{li2007local} which exploited geodesic distance maps derived from minimal weighted curve length to generate Voronoi regions and to reconstruct the respective Voronoi interface. The region-based homogeneity features and the edge appearance features can be simultaneously encoded into Voronoi diagram construction. The main advantages of this model lie at the efficiency of the numerical solutions and at the flexibility in the extension to multi-region segmentation applications.  However, the isotropy limitation of the metrics used in this classical model may suffer from the shortcut problem, i.e. the contour evolution stabilizes before reaching the true boundaries,  especially when the object regions have complicate intensity distributions.

\noindent\emph{Contributions and Paper Outline}.
In this paper, we propose a new Voronoi diagram-based contour evolution model based on a Finsler metric with an asymmetric quadratic form, which is capable of integrating asymmetry enhancement as well as the image features for Voronoi diagram construction.  The main contributions are twofold:
\begin{itemize}
\item Firstly, we generalize Li and Yezzi's isotropic dual-front model~\cite{li2007local} to an asymmetry-enhanced case. Instead of using direction-independent  metrics, the proposed model encapsulates a type of Finsler geodesic metric with an asymmetric quadratic form. The asymmetry property of the considered metric serves as an efficient constraint for front propagation, so as to reduce the risk of the shortcut issue in Voronoi diagram-based segmentation applications.
\item Secondly, we propose a new method for the construction of data-driven asymmetric quadratic metrics. The ingredients of these metrics  are respectively derived from the anisotropic image edge-based features such as image gradients, a variety of region-based homogeneity terms, and the predicted motion directions of the contour. As a result, the image segmentation via the proposed dual-front model can blend the benefits from these features.
\end{itemize}

The structure of this manuscript is organized as follows. In Section.~\ref{sec_VorDiagram}, we introduce the preliminaries on the construction of the Voronoi diagram and the corresponding applications for dual-front models. Sections~\ref{sec_MetricConstruct} and~\ref{Sec:Numerical} present the main contribution of this paper: the construction of data-driven asymmetric quadratic metrics for Voronoi diagram-based image segmentation.  The experimental results on both synthetic and real images are given in Section \ref{sec_Experiment} and the conclusion is presented in Section~\ref{sec_Conclusion}. The current document is an extension of the short conference paper presented in~\cite{chen2018asymmetric}, upon which more contributions were added.

\section{Voronoi Diagram-based Contour Evolution}
\label{sec_VorDiagram}

\subsection{Preliminaries on Voronoi Diagram}
\label{subsec_Voronoi}
The Voronoi diagram is known as a powerful geometric tool for domain partitioning and image segmentation~\cite{peyre2006geodesic,bougleux2008anisotropic,leibon2000delaunay,peyre2010geodesic}. The construction of Voronoi diagram, regarded as a tessellation of the domain $\Omega\subset\bR^d$, can be naturally and efficiently implemented through $n$ geodesic distance maps. Each geodesic distance map, denoted by $\rD_i:\Omega\to\bR^+_0$ and indexed by $i=1,2,\cdots,n$, is constructed from a set $\cS_i\subset\Omega$ of source points.  The distance value $\rD_i(x)$ at any point $x\in\Omega\backslash\cS_i$ represents the weighted length of the minimal path between the source point set $\cS_i$ and $x$, measured by a local metric $\cF:\Omega\times\bR^d\to\bR^+$. At any fixed point $x\in\Omega$, a metric $\cF(x,\cdot)$ can be defined using an asymmetric norm $F_x(\cdot)$ on $\bR^d$ such that $\cF(x,\fu)=F_x(\fu)$ for any vector $\fu\in\bR^d$.

Let us denote by $\Lip_{x,y}$ the set of Lipschitz continuous curves $\gamma:u\in[0,1]\mapsto\gamma(u)\in\Omega$ subject to $\gamma(0)=x$ and $\gamma(1)=y$. Once the geodesic metric $\cF$ and the source point set $\cS_i$ are given, a crucial ingredient for estimating a geodesic distance map lies at the definition of the minimal weighted length  $\kD_\cF(x,y)$
\begin{equation}
\label{eq:MinLength}
\kD_\cF(x,y):=\inf_{\gamma\in\Lip_{x,y}}\left\{\int_0^1\cF(\gamma(u),\gamma^\prime(u))du\right\},
\end{equation}
where $\gamma^\prime(u)$ is the first-order derivative of curve $\gamma$.
Then the geodesic distance map $\rD_i$ associated to the set $\cS_i$ reads 
\begin{equation}
\label{eq:Distance}
\rD_i(x)=\min_{y\in\cS_i}\kD_\cF(y,x).	
\end{equation}
As in~\cite{mirebeau2014anisotropic,mirebeau2014efficient,mirebeau2019riemannian}, the geodesic distance  map $\rD_i$ is a viscosity solution to the Eikonal equation  
\begin{equation}
\label{eq:EikonalPDE}
\begin{cases}
\cH(x,\nabla\rD_i(x))=\frac{1}{2},\quad &\forall x\in\Omega\backslash\cS_i,\\
\rD_{i}(x)=0,&\forall x\in\cS_i	
\end{cases}
\end{equation}
where  $\nabla\rD_i$ is the standard Euclidean gradient of $\rD_i$ over the domain $\Omega\backslash\cS_i$, and $\cH:\Omega\times\bR^d\to[0,\infty]$ is a Hamiltonian
\begin{equation}
\label{eq:Hamiltonian}
\cH(x,\fu):=\sup_{\fv\in\bR^d}\left\{\<\fu,\fv\>-\frac{1}{2}\cF(x,\fv)^2\right\}.	
\end{equation}

From the geodesic distance maps $\rD_i$ for $1\leq i\leq n$, one can generate $n$ Voronoi regions, denoted by $\Vor(\cS_i)\subset \Omega$, each of which is defined  as follows
\begin{equation}
\label{eq:VoronoiRegion}
\Vor(\cS_i)=\{x\in\Omega;\rD_i(x)<\rD_j(x),\,\forall j\neq i\}.
\end{equation}
In this case, a point $x\in\Vor(\cS_i)$ implies that $x$ is precisely closer to $\cS_i$ than to any other $\cS_j$ ($\forall j\neq i$) in the sense of geodesic distance. We say that two distinct Voronoi regions $\Vor(\cS_i)$ and $\Vor(\cS_j)$ are adjacent if the set of points which are equidistant to $\cS_i$ and $\cS_j$ is not empty.

\noindent\emph{Voronoi diagram associated to multiple metrics}. We have presented the basic procedure for the construction of Voronoi diagram based on the geodesic distance maps. One can see that all the geodesic distance maps $\rD_i$ are derived using the same metric $\cF$. In our Voronoi diagram-based contour evolution model (also in~\cite{li2007local}), we allow each geodesic distance map $\rD_i$ to be computed using different metrics $\cF_i$
\begin{equation}
\rD_i(x)=\min_{y\in\cS_i}\kD_{\cF_i}(y,x).
\end{equation}
In other words, each geodesic distance map $\rD_i$ is estimated by an individual metric $\cF_i$.

\subsection{Voronoi Diagram for Dual-front Models}
\label{subsec:IDF}
In this section, we briefly summarize the Voronoi diagram-based dual-front models, by which the image segmentation problems can be naturally addressed in an iterative manner. Significant examples include the model based on the VII scheme~\cite{dubrovina2015multi} and the model based on geodesic distance~\cite{li2007local}. In each iteration, both dual-front models can be loosely divided into three steps: extracting the offset lines of the evolving contour, building the Voronoi diagram and reconstructing the new contour. The generic algorithm for the Voronoi diagram-based dual-front models are presented in Algorithm~\ref{algo:Dual-Front}. 

Let $\Gamma\subset\Omega$ be a finite set of curves, which is referred to as  a contour in the following. During the contour evolution, $\Gamma$ can be taken as the input of both dual-front models for 2D image segmentation\footnote{In this paper, we focus on the 2D image segmentation and the extension to 3D volume segmentation is straightforward.}, i.e. the dimension $d=2$.  In general, the input $\Gamma$ partitions the image domain $\Omega$ into $n$ open and bounded connected regions $\cR_i$ for $1\leq i\leq n$ such that $\Omega=\cup^n_i\cR_i\cup\Gamma$ and $\Gamma=\cup_i^n\partial\cR_i$. Let $\Gamma_{i,j}=\partial\cR_i\cap\partial\cR_j$ be the interface between two adjacent regions $\cR_i$ and $\cR_j$. We consider a collection of subregions $\Vor(\Gamma_{i,j})\subset\Omega$ 
\begin{equation}
\label{eq:VoronoiBound}
\Vor(\Gamma_{i,j})=\left\{x\in\Omega;\rE(x;\Gamma_{i,j})<\rE(x;\Gamma\backslash\Gamma_{i,j}) \right\},
\end{equation}
where $\rE(x;S)$ denotes the unsigned Euclidean distance between a point $x$ and a set $S$, i.e.
\begin{equation}
\label{eq_EuclideanDist}
\rE(x;S):=\min_{y\in S}\|x-y\|.
\end{equation}

\begin{algorithm}[!t]
\caption{Voronoi Diagram-based Dual-front Model}	
\label{algo:Dual-Front}
\begin{algorithmic}
\renewcommand{\algorithmicrequire}{\textbf{Input:}}
\renewcommand{\algorithmicensure}{\textbf{Output:}}
\Require Initial contour $\Gamma$.
\Ensure New contour  $\Gamma^*$.
\end{algorithmic}
\begin{algorithmic}[1]
\While{\emph{Stopping criteria are not satisfied}}
\State Extract all the offset lines $\cC^\ell_i,\,1\leq i\leq n$ of $\Gamma$.
\State Build the Voronoi diagram with respect to the offset lines $\cC_i$.
\State Reconstruct a new contour $\Gamma^*$ as the collection of the interfaces between each pair of adjacent Voronoi regions.
\State Set $\Gamma\gets\Gamma^*$.
\EndWhile
\end{algorithmic}
\end{algorithm}

\subsubsection{The VII scheme-based dual-front model}
In the VII scheme-based dual-front model~\cite{dubrovina2015multi}, the offset lines are extracted from an evolved LSF $\phi$. The evolution of $\phi$ is driven by minimizing an energy functional comprised of  an image data term and a regularization term. Significant examples considered in~\cite{dubrovina2015multi} involve the region competition model~\cite{zhu1996region} and the pairwise similarity models~\cite{jung2012nonlocal,bertelli2008variational}, of which the image data-driven terms can be summarized as
\begin{equation}
\label{eq:Energy}
E_{\rm data}(\Gamma)=\sum_{i=1}^n E_i(\Gamma_i),
\end{equation}
where $\Gamma_i:=\cup_j\Gamma_{i,j}$. The motion equation with respect to a time parameter $t$ for minimizing $E_{\rm data}$ can be formulated for any point $x\in\Gamma$ as follows
\begin{equation}
\label{eq:CMotion}
\frac{\partial\Gamma}{\partial t}=-\frac{\partial E_{\rm data}}{\partial\Gamma}=\sum_{i\in\rA(x)}\tilde\xi_i(x)\cN_i(x),
\end{equation}
where $\cN_i(x)$ stands for the inward unit normal to the boundary $\partial\cR_i$ at $x$, and $\rA(x)$ is a set involving all indices $i$ such that $x\in\partial\cR_i$. The velocity functions $\tilde\xi_i$ for $1\leq i\leq n$ are defined being such that 
\begin{equation}
\label{eq:SingularVelocity}
\frac{\partial E_i}{\partial\Gamma_i}=	-\tilde\xi_i(x)\cN_i(x),~\forall x\in\Gamma_i.
\end{equation}
As discussed in~\cite{dubrovina2015multi}, these velocity functions $\tilde\xi_i$ should be extended to the image domain $\Omega$ or to a narrow band neighbourhood of $\Gamma$, in order to drive the update scheme for the LSF $\phi$. In the following, we denote by $\xi_i$ the respective extension of $\tilde\xi_i$ for $1\leq i\leq n$, subject to $\xi_i(x)=\tilde\xi_i(x),\,\forall x\in\Gamma_i$.

The VII method initializes the evolving LSF $\phi$ as $\forall x\in\Omega,\,\phi(x,0)=\rE_\Gamma(x)$ with $\rE_\Gamma(x):=\rE(x;\Gamma)$ being the unsigned Euclidean distance map associated to $\Gamma$, see Eq.~\eqref{eq_EuclideanDist}. Using the extended velocity functions $\xi_i$,  one can define a new velocity function $\xi_{\rm ext}$ formulated as follows~\cite{dubrovina2015multi}
\begin{equation}
\label{eq:ExtVelocity}
\xi_{\rm ext}(x)	=\xi_j(x)-\xi_i(x),\quad \forall x\in\cR_i\cap\Vor(\Gamma_{i,j}).
\end{equation}
Accordingly, the LSF evolution equation associated to  Eq.~\eqref{eq:CMotion} reads as
\begin{equation}
\label{eq:LSE}
\frac{\partial\phi}{\partial t}=\xi_{\rm ext}\|\nabla\phi\|.
\end{equation}

At some time $t>0$, each evolved offset line, noted as $\tilde\cC^\ell_i$, is the $\ell$-level set of the solution $\phi(x,t)$ to the evolution equation~\eqref{eq:LSE}. For a small $\ell\in\bR^+$, one has
\begin{equation*}
\tilde\cC^\ell_i:=\{x\in\cR_i;\phi(x,t)=\ell\}.
\end{equation*}

Once all the offset lines $\{\tilde\cC^\ell_{i}\}_{1\leq i\leq n}$ are extracted, the construction of the Voronoi diagram in a narrow band $\tilde U=\{x\in\Omega;\phi(x,t)<\ell\}$ can be implemented using the method presented in Section~\ref{subsec_Voronoi},  by setting $\cS_i:=\tilde\cC^\ell_i$ and $\cF(x,\fu)=\|\fu\|,\,\forall x\in \tilde U$.

\subsubsection{The geodesic distance-based dual-front model}
Li and Yezzi~\cite{li2007local} proposed a dual-front model, where the construction of the Voronoi diagram is implemented by geodesic distances associated to a family of data-driven isotropic metrics. In the basic setting of the Li-Yezzi model, the offset lines $\cC_i^\ell$ of the boundaries $\Gamma_i:=\partial\cR_i$ ($1\leq i\leq n$) can be simply extracted by leveraging the $\ell$-level set of $\rE(x,\Gamma_i)$
\begin{equation}
\label{eq:OffsetLines}
\cC^\ell_i:=\big\{x\in\cR_i;\rE(x,\Gamma_i)=\ell\big\}.	
\end{equation}

The second step in the Li-Yezzi model is to reconstruct the Voronoi regions $\Vor(\cC_i^\ell)$ within a neighbourhood  $U_{\Gamma}$ of $\Gamma$ 
\begin{equation}
\label{eq:TubeNeigh}
U_\Gamma:=\big\{x\in\Omega;\rE_\Gamma(x)<\ell\big\}.
\end{equation}
via a family of geodesic distance maps $\rD_i$, as in Section~\ref{subsec_Voronoi}. Each geodesic distance map $\rD_i$ exploits the offset line $\cC^\ell_i$ as the  set of source points, and can be estimated by solving the Eikonal PDE~\eqref{eq:EikonalPDE}. In the Li-Yezzi model, the distance maps $\rD_i$ are estimated using isotropic Riemannian metrics. These metrics integrate both the magnitude of image gradients and the mean and variance of the image intensities in each region $\cR_i$. As an important shortcoming, the isotropic metrics invoked in the Li-Yezzi dual-front model are independent to the expected motion directions of the contour, which may increase the possibility of the evolving contour to suffer from the shortcut problem. In this paper, we propose a new dual-front model to overcome this drawback, by extending the isotropic metrics to the asymmetric quadratic metrics.

\section{Voronoi Diagram for Dual-front Model from Asymmetric Quadratic Metrics}
\label{sec_MetricConstruct}
In this section, we present our core contribution on the construction of the asymmetric geodesic metrics depending on the image data such as the region-based homogeneity terms and image gradients. We first introduce the general form of these metrics considered and then present the principle for their construction in the context of contour evolution.

\subsection{Asymmetric Quadratic Metrics}
Let $\bS^+_2$ be a set of symmetric positive definite  matrices of size $2\times 2$. In this section, we consider an asymmetric quadratic metric which is made up of a tensor field $\cM:\Omega\to\bS^+_2$ and a vector field $\omega:\Omega\to\bR^2$
\begin{equation}
\label{eq:AQMetric}
\cF(x,\fu):=\sqrt{\langle \fu,\cM(x)\fu\rangle+\<\fu,\omega(x)\>_-^2}, 
\end{equation} 
where  $\<\cdot,\cdot\>$ stands for the Euclidean scalar product on $\bR^2$. The second term in Eq.~\eqref{eq:AQMetric} involves a scalar product which can be expressed as~\cite{duits2018optimal}  
\begin{equation}
\label{eq:AsyQuadNorm}
\<\fu,\fv\>_-:=\max\{-\<\fu,\fv\>,0\},\quad \forall \fu,\,\fv\in\bR^2.
\end{equation}
and $\<\fu,\fv\>_-^2:=(\max\{-\<\fu,\fv\>,0\})^2$.

The asymmetric quadratic metric $\cF$ in Eq.~\eqref{eq:AsyQuadNorm} should be positive, $1$-homogeneous and convex w.r.t its second argument. The positivity and homogeneity properties are clearly satisfied. The proof for the convexity property of the asymmetric quadratic metric $\cF$ is presented in Proposition~\ref{prop:Convexity} of Appendix~\ref{subsec_Convexity}. 
 Moreover, the metric $\cF$ formulated in Eq.~\eqref{eq:AQMetric} is asymmetric with respect to its second argument due to the existence of the second term~\eqref{eq:AsyQuadNorm}. Note that when the vector field $\omega\equiv\mathbf 0$, the metric $\cF$ gets to a symmetric Riemannian metric, i.e. $\cF(x,\fu)=\sqrt{\langle \fu,\cM(x)\fu\rangle}$. 

\begin{figure}[t]
\centering
\subfigure[]{\includegraphics[width=2.8cm]{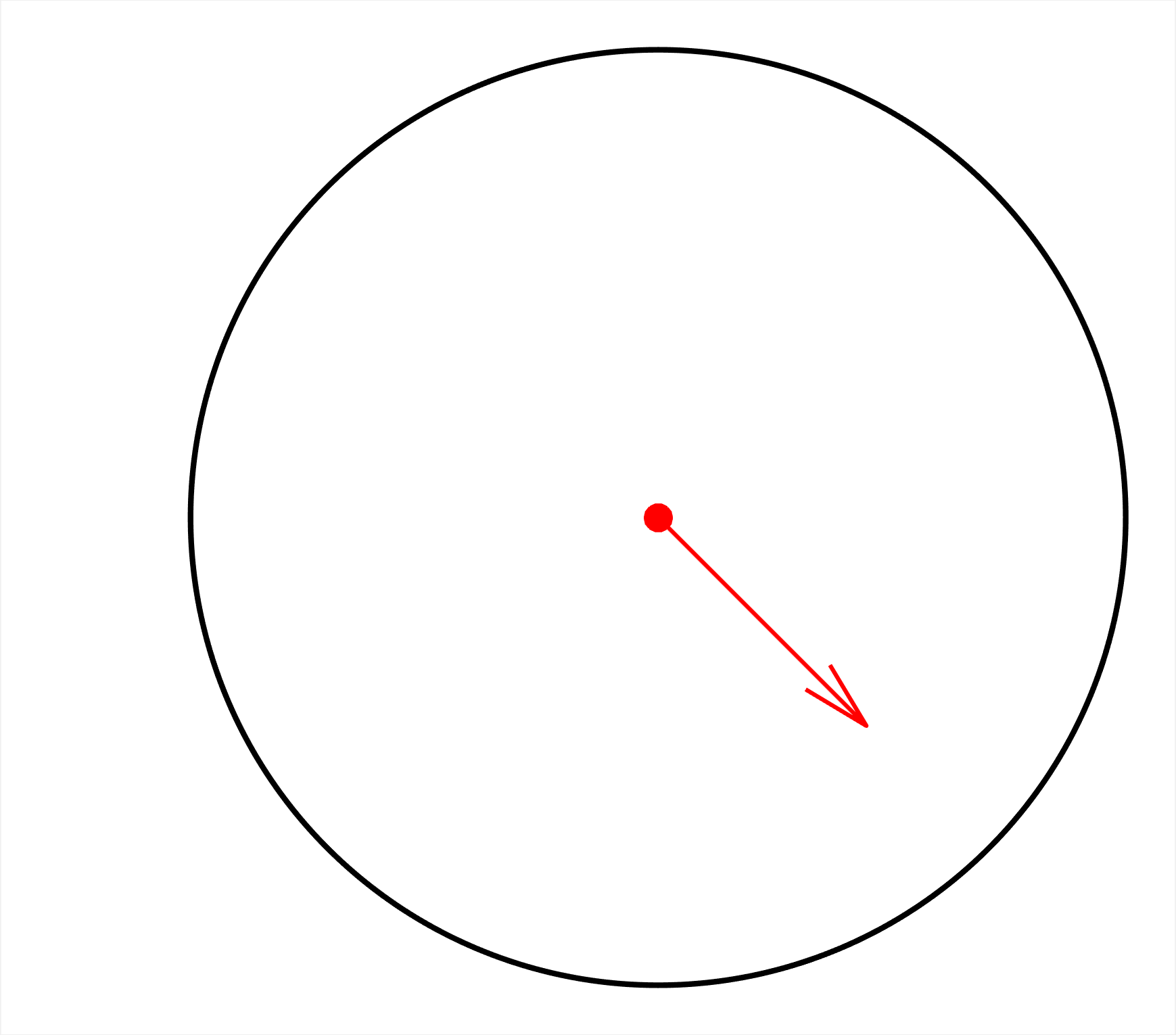}}~\subfigure[]{\includegraphics[width=2.8cm]{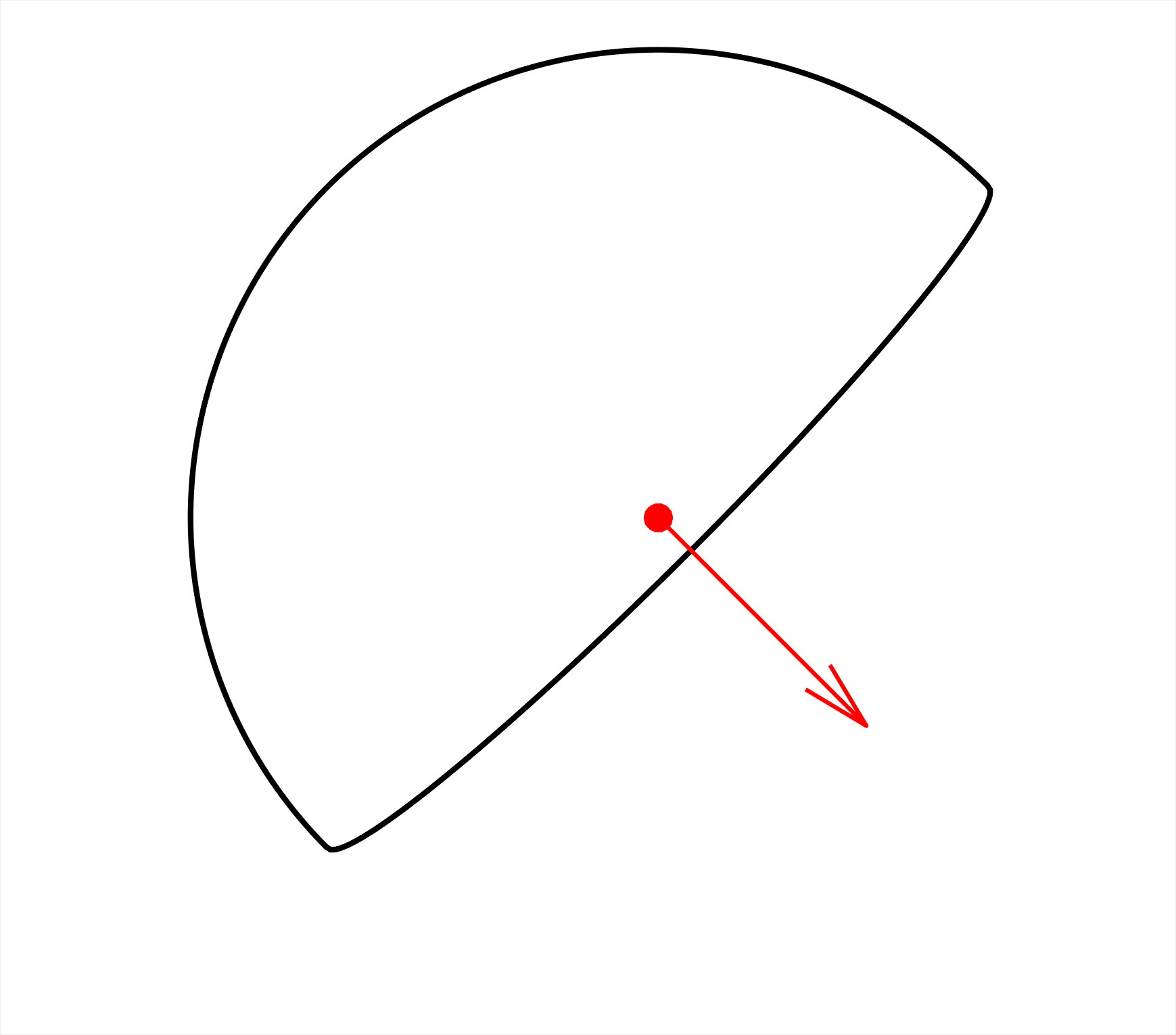}}~\subfigure[]{\includegraphics[width=2.8cm]{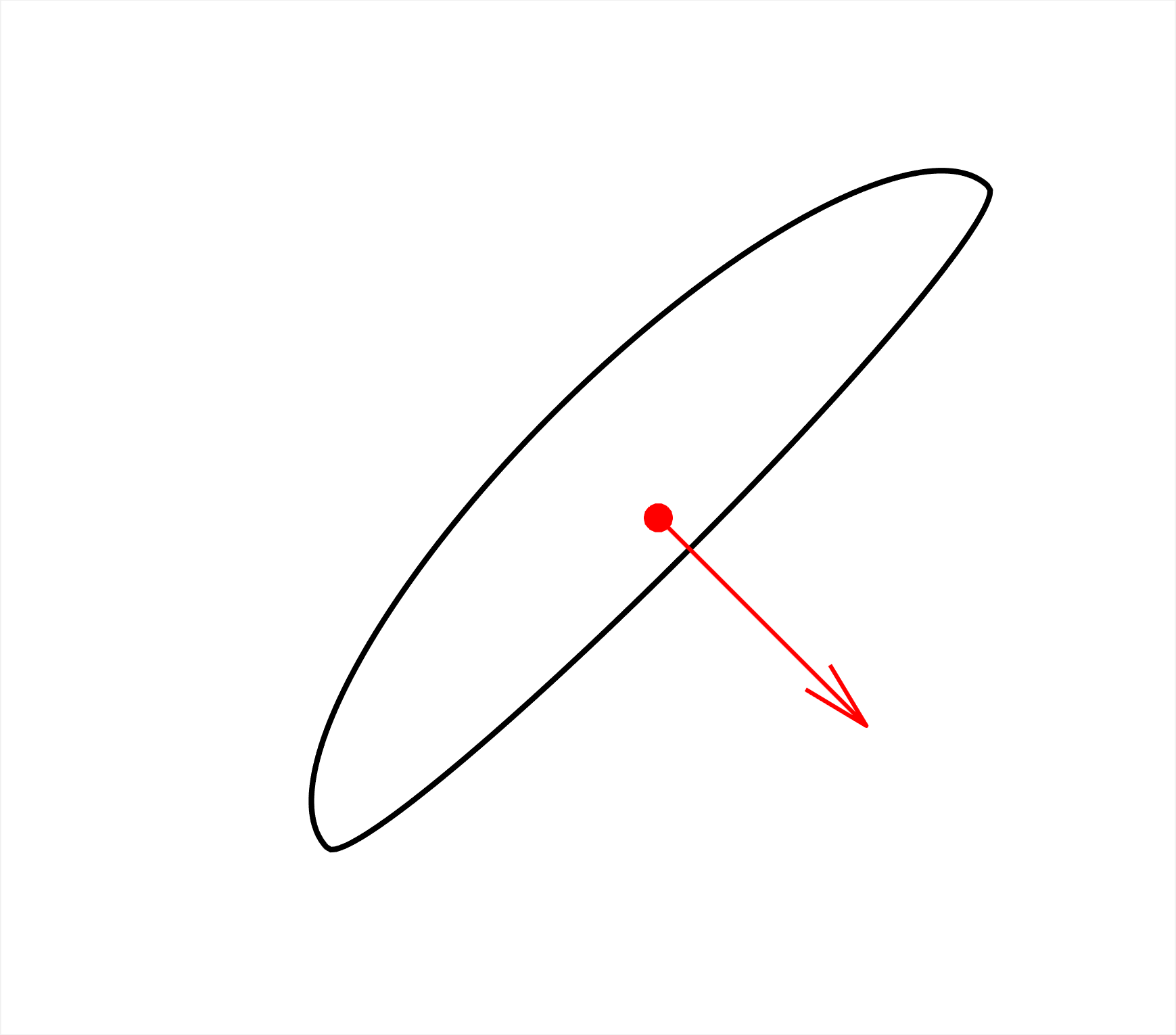}}
\caption{Unit balls $\cB$ for the metrics $\cF$ with respect to different tensor fields $\cM$ and vector fields $\omega$, see text. The black lines are the boundaries of $\cB$, the red dots are the origin and the red arrows indicate the orientation $\fp=(1,-1)^T$}
\label{fig_Balls}
\end{figure}

\begin{figure*}[t]
\centering
\includegraphics[width=17.5cm]{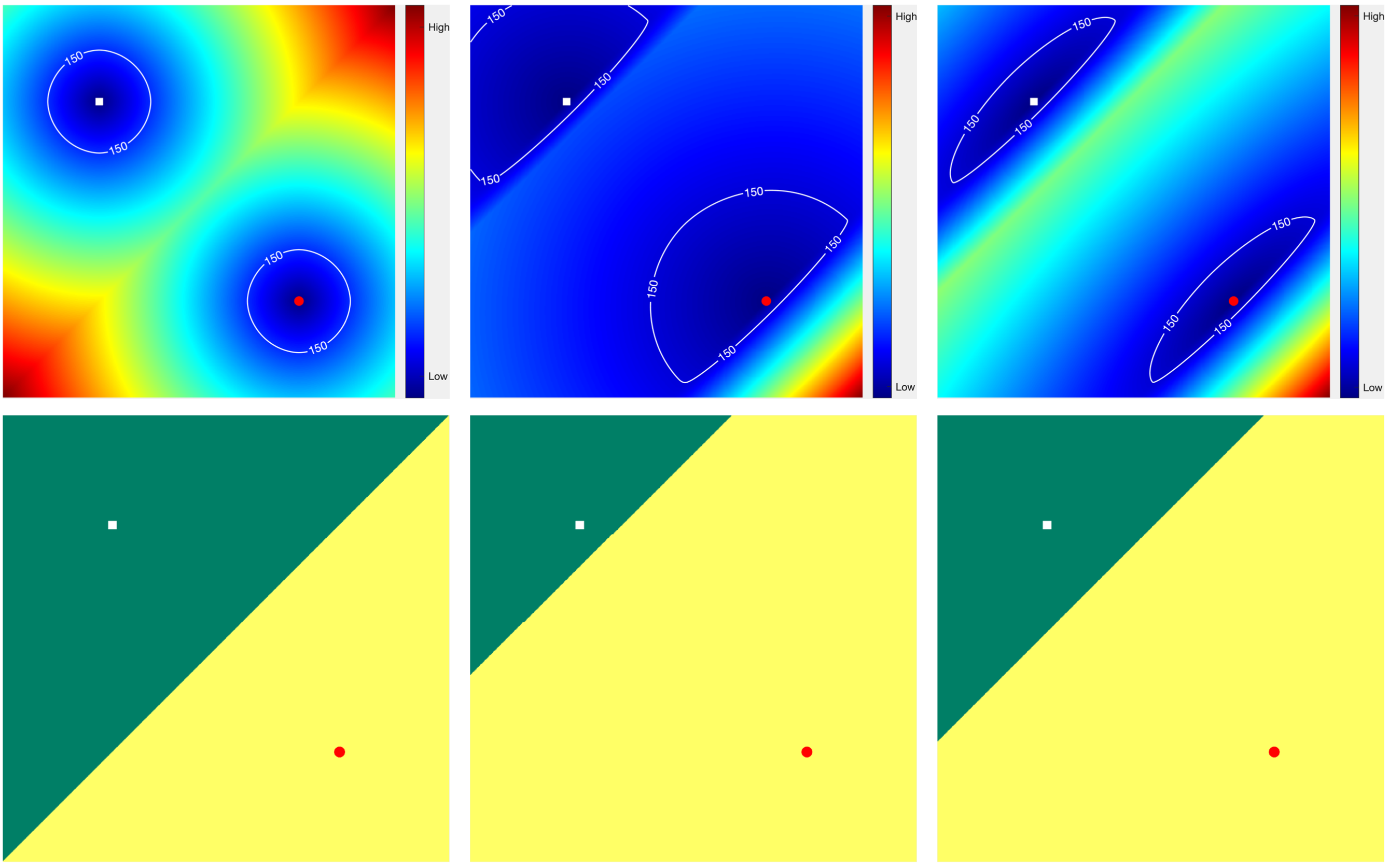}
\caption{Geodesic distance maps (first row) and the associated Voronoi regions (second row) with respect to different geodesic metrics. The red dots and white squares represent source points for different Voronoi regions. The white lines represent the $150$-level set lines of the geodesic distance maps. \textbf{Columns} 1-3: The geodesic distance maps and the corresponding Voronoi regions are computed using geodesic metrics for which the control sets are shown in  Figs.~\ref{fig_Balls}a to~\ref{fig_Balls}c, respectively}
\label{fig_LS}
\end{figure*}

It is a popular way to utilize the tool of  control sets for the visualization of  a geodesic metric. The control set $\cB(x)$ for any point $x$ is defined as the unit ball of the metric $\cF(x,\cdot)$
\begin{equation}
\label{eq:ControlSet}
\cB(x)=\{\fu\in\bR^d;\cF(x,\fu)\leq 1\}.	
\end{equation}
The unit ball $\cB(x)$ is governed by both of the matrix $\cM(x)$ and the vector $\omega(x)$. In Fig.~\ref{fig_Balls}, we illustrate the unit balls $\cB$ with respect to different matrices $\cM(x)$ and vectors $\omega(x)$. In Fig.~\ref{fig_Balls}a, we set $\cM(x)=\Id$ where $\Id$ is an identity of size $2\times2$, and $\omega(x)=\mathbf{0}$. The corresponding unit ball appears to be a disk, since $\cF$ gets to be isotropic in this case. Furthermore, assuming that $\fp=(1,-1)^T$ and keeping $\cM(x)=\Id$, the vector $\omega(x)=10\,\fp$ leads to a unit ball close to a half disk, as depicted in Fig.~\ref{fig_Balls}b. The metric $\cF(x,\fu)$ has high values if the vectors $\langle\fu,-\fp\rangle\approx\|\fu\|$. Finally,  we use $\omega(x)=10\,\fp$ and 
$\cM(x)=10\fp\fp^T+\fp_\perp\fp_\perp^T$, where $\fp_\perp$ is the vector orthogonal to $\fp$. In this case, the control set $\cB(x)$ approximates a half ellipse, as shown in Fig.~\ref{fig_Balls}c.
In Fig.~\ref{fig_LS}, we illustrate the geodesic distance maps and the corresponding Voronoi regions. In each column of this figure, the distance map is estimated using a metric $\cF(x,\cdot)$ at any point $x$ satisfying the control set shown in the respective column of Fig.~\ref{fig_Balls}.

\subsection{Constructing Data-driven Asymmetric Quadratic Metrics}
\label{subsec_RegionalMetric}
 
\subsubsection{Data-driven Asymmetric Quadratic Metrics}
We denote by $\Gamma$ a given contour, which partitions the image domain $\Omega$ into $n$ open and bounded regions $\cR_i$ and  yields $n$ offset lines $\cC^\ell_{i}$ by means of Eq.~\eqref{eq:OffsetLines}. In order to reconstruct the Voronoi regions $\Vor(\cC^\ell_i)$, we consider the following image data-driven asymmetric quadratic metrics 
\begin{equation}
\label{eq:DataMetric}
\kF_i(x,\fu)=\psi_i(x)\sqrt{\langle \fu,\cM(x)\fu\rangle+\<\fu,\omega_i(x)\>_-^2},
\end{equation}
where $\psi_i$ is a positive scalar-valued function.
The tensor field $\cM$ carries out the edge anisotropy and appearance features derived from the image gradients. The scalar-valued functions $\psi_i$ and the vector fields $\omega_i$ with $1\leq i\leq n$ are dependent to the region-based homogeneity terms considered. The computation for these ingredients of the data-driven asymmetric quadratic metrics~\eqref{eq:DataMetric} will be described in the following.

In the proposed dual-front model, in order to reduce the computation costs, the estimation for each geodesic distance map $\rD_i$ is restricted in a tubular neighbourhood $U_{\Gamma_i}\subset U_\Gamma$ of the boundary $\Gamma_i=\partial\cR_i$, where $U_\Gamma$ is the neighbourhood of the whole contour $\Gamma$, see Eq.~\eqref{eq:TubeNeigh}. Such a neighbourhood $U_{\Gamma_i}$, abbreviated as $U_i$, can be expressed as 
\begin{equation}
\label{eq:NeighBoundary}
U_i:=\left\{x\in U_\Gamma;\min_{y\in\partial\Gamma_i}\|x-y\|<\ell\right\}.	
\end{equation}
Furthermore, we also apply that restriction to the construction of the asymmetric quadratic metrics~\eqref{eq:DataMetric}. In other words,  each data-driven metric $\kF_i$ is defined over the domain $ U_i\times\bR^2$.   

\subsubsection{Principle for constructing data-driven metrics $\kF_i$}

Given a suitable constant $\epsilon\in\bR^+_0$, one can extract the $\epsilon$-level set lines $\{\cC_i^\epsilon\}_i$ from the Euclidean distance map $\rE_\Gamma(x)$. In the VII scheme-based active contour model~\cite{dubrovina2015multi}, the LSF evolution equation~\eqref{eq:LSE} also characterizes the evolution of the level set lines $\cC_i^\epsilon$ for $\epsilon>0$, leading to a fact that a point $x\in \cC_i^\epsilon$ will move along the direction $-\xi_{\rm ext}(x)\nabla\rE_\Gamma(x)$, where $\xi_{\rm ext}$ is the velocity function defined in Eq.~\eqref{eq:ExtVelocity}. 
In this case, we introduce a family of vector fields $\fn_i:U_i\to\bR^2$ indexed by $1\le i\leq n$ as follows
\begin{align}
\label{eq:weightedNormals}
\fn_i(x)=
\begin{cases}
\sign(-\xi_{\rm ext}(x))\nabla\rE_\Gamma(x),&\forall x\in U_i\backslash\Gamma\\
\sign(\xi_{i}(x)-\xi_j(x))\cN_i(x),&\forall x\in\Gamma_i\cap\Gamma_j,
\end{cases}	
\end{align}
and $\fn_i(x)=\mathbf{0}$ otherwise, 
where $\sign(a)$ is the sign of a scalar value $a \in \bR$. Recall that $\cN_i(x)$ is the inward  normal vector to the boundary $\Gamma_i$ at $x$. The vectors $\fn_i(x)$ indicate the motion directions of the level set $\cC^\epsilon_i$ for $\epsilon\geq 0$, which point to the desired boundary from $x$. Nevertheless, we consider to exploit these vector fields $\fn_i$ to construct the vector fields $\omega_i$ for $1 \leq i\leq n$.

From the viewpoint of front propagation, a front $\zeta_i^\epsilon:=\{x;\rD_i(x)=\epsilon\}$ is a level set line of the geodesic distance map $\rD_i$ emanating from the offset line $\cC^\ell_i$. At some point $x$, the advancing direction, denoted by $\kN_i(x)\in\bR^2$, of the front $\zeta_i^\epsilon$ subject to $\rD_i(x)=\epsilon$ is positively proportional to $\nabla\rD_i(x)$. Therefore, we encourage that the front $\zeta_i^\epsilon$, which passes through the point $x$, propagates fast in case its advancing direction $\kN_i(x)$ forms an acute angle with $\fn_i(x)$, i.e. $\langle\kN_i(x),\fn_i(x) \rangle>0$.  Towards this purpose, we construct the vector fields $\omega_i$ for $1 \leq i \leq n$ as
\begin{equation}
\label{eq:omega}
\omega_i(x):=\mu\fn_i(x),
\end{equation}
where $\mu\in\bR^+$ is a constant as a weighted parameter. 

In addition, the function $\psi_i$ can be estimated for any point $x\in U_i\cap\Vor(\Gamma_{i,j})$ as follows
\begin{equation}
\label{eq:PSI}
\psi_i(x):=\exp\left(\frac{\alpha\,(\xi_i(x)-\xi_j(x))}{\displaystyle\sup_{y}|\xi_i(y)-\xi_j(y)|}\right), 
\end{equation}
where $\alpha\in\bR^+$ is a constant and $\Vor(\Gamma_{i,j})$ is the Voronoi region associated to $\Gamma_{i,j}$ as defined in Eq.~\eqref{eq:VoronoiBound}.

\noindent\emph{Remark}.
Note that the vector fields $\fn_i$ (or $\omega_i$) dominate the front propagation speed especially within the homogeneous regions where the image gradients are small. As a consequence, the use of asymmetric quadratic metrics~\eqref{eq:DataMetric} allows us to perform the front propagation independently to the weighted functions $\psi_i$. Therefore, in contrast to the classical isotropic dual-front model~\cite{li2007local}, the introduced dual-front evolution scheme featuring asymmetric property could in principle be implemented using one single metric with $\psi_i\equiv1$ for $1\leq i\leq n$ such that $\kF(x,\fu)=\kF_i(x,\fu)$ for any point $x\in U_\Gamma\backslash\Gamma$, and $\kF(x,\fu)=\sqrt{\langle\fu,\cM(x)\fu \rangle},\,\forall x\in\Gamma$.
This potential simplification was one of our initial motivations for the study of Voronoi diagram-based active contours associated to asymmetric metrics. However, in the end, we found that the best efficiency can be achieved by combining the advantages of both (i) asymmetric geodesic metrics, and (ii) distinct weighted functions $\psi_i$ for the propagation of the respective fronts.	

\subsection{Edge Anisotropy Features from Image Gradients}
\label{subsec:EdgeAniso}
The construction of the tensor field $\cM$ relies on the image gradients, which carries out the edge anisotropy information.  With respect to  a  vector-valued image $\fI=(I_1,I_2,I_3):\Omega\to\bR^3$ in the RGB color space, we apply the method introduced in~\cite{di1986note,sochen1998vision} to estimate the image gradients for a Gaussian-smoothed image.
Let $G_\sigma$ be a Gaussian kernel with a standard deviation $\sigma$ and we denote by $\nabla G_\sigma$ the Euclidean  gradient of $G_\sigma$. At each point $x$, we first compute a Jacobian matrix $\cW(x)$ of size $2\times 3$
\begin{equation}
\label{eq:Jacobian}
\cW(x)=
\begin{pmatrix}
\nabla G_\sigma\ast I_1,\nabla G_\sigma\ast I_2,\nabla G_\sigma\ast I_3
\end{pmatrix}(x),
\end{equation}
where and `$\ast$' is a convolution operator. Furthermore, for a gray level image $I:\Omega\to\bR$, the equation~\eqref{eq:Jacobian} gets to be $\cW(x)=(\nabla G_\sigma\ast I)(x)$. When smoothing the images $\fI$  via $G_\sigma$, high values of $\sigma$ can suppress the effects from noise, but may potentially increase the risk of missing weak edges. 

The eigenvectors of the matrix $\cW(x)\cW(x)^T$, referred to as $\vartheta_k(x)\in\bR^2$ for $k=1,2$, can be used to characterize the edge anisotropy features. Among them, the eigenvector $\vartheta_1(x)$ which corresponds to the largest eigenvalue of $\cW(x)\cW(x)^T$ can be used to indicate the direction perpendicular to the edge tangent at $x$.

The edge appearance features are carried  by a scalar-valued function $\eta:\Omega\to[0,1]$ defined by
\begin{equation}
\label{eq_EdgeAppearance}
\eta(x)=\frac{\|\cW(x)\|_{\rm F}}{\sup_y\,\|\cW(y)\|_{\rm F}},\quad \forall x\in\Omega
\end{equation}
where $\|\cW(x)\|_{\rm F}$ is the Frobenius norm of the matrix $\cW(x)$
\begin{equation*}
\|\cW(x)\|_{\rm F}^2=\sum_{m=1}^3\|(\nabla G_\sigma\ast I_m)(x)\|^2.
\end{equation*}

By means of the eigenvectors $\lambda_1$, $\lambda_2$ and the normalized  Frobenius norms $\eta$, we construct the tensor field $\cM$ as follows
\begin{equation}
\label{eq:Eigen}
\cM(x)=\sum_{k=1}^{2}\lambda_k(x)\vartheta_k(x)	\vartheta_k(x)^T,~\text{~if~}\eta(x)\neq 0,
\end{equation}
and $\cM(x)=\Id$ otherwise, where $\Id$ is the identity of size $2\times 2$ and where $\lambda_1,\,\lambda_2:\Omega\to\bR^+$ reads
\begin{align}
\label{eq:Eigens1}
&\lambda_1(x)=\exp\big((\beta+\rho)\,\eta(x)\big),\\
\label{eq:Eigens2}
&\lambda_2(x)=\exp\big(\rho\,\eta(x)\big),
\end{align}
where $\beta,\,\rho\in\bR^+$ are two constants. Specifically, the values of $\beta$  dominate the anisotropy property of $\cM$, while $\rho$ controls the relative importance on the edge appearance features. 

Eventually, we smooth the tensor field $\cM$ via a Gaussian kernel $G_q$  with standard deviation $q$
\begin{equation}
\label{eq:SmoothedEdgeTensor}
\tilde\cM(x)=(G_q\ast\cM)(x),\quad \forall x\in\Omega.	
\end{equation}
Each entry of the matrix $\tilde\cM(x)$ is generated by convolving the corresponding entry of $\cM(x)$ via $G_q$. The value of the standard deviation $q=2$ is fixed in the following experiments.

\begin{figure*}[t]
\centering
\includegraphics[height=4.5cm]{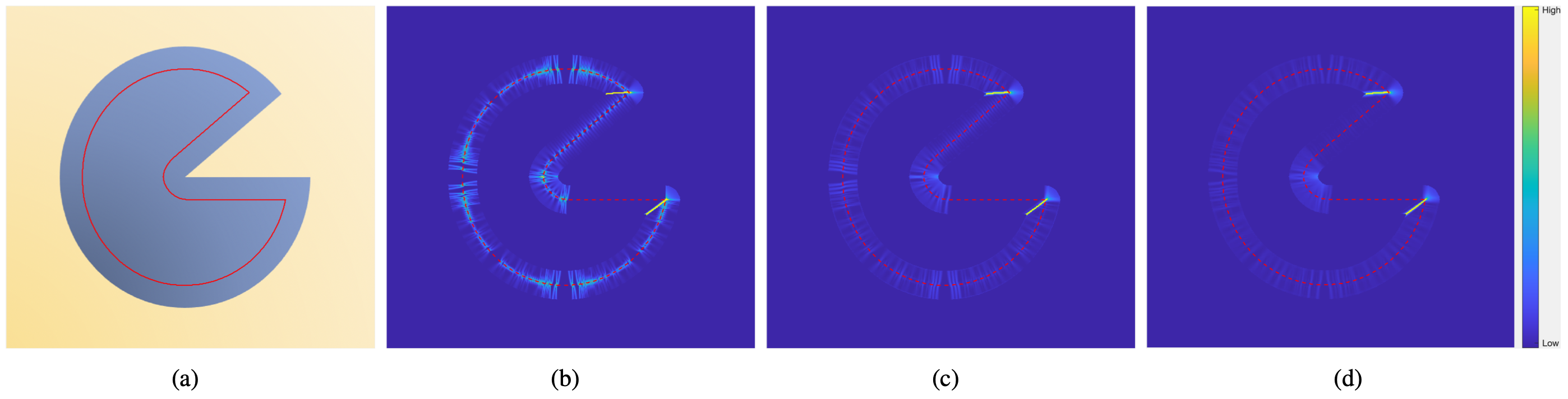}
\caption{Visualization for Frobenius norms of Jacobian matrices of the vector fields $\fn_i$ and $\tilde\fn_i$. (\textbf{a}) A synthetic image with an input contour $\Gamma$ denoted by a red solid line. (\textbf{b}) Visualization for the Frobenius norms $\|\nabla\fn_i\|_{\rm F}$. (\textbf{c}) and (\textbf{d}) Visualization for the Frobenius norms $\|\nabla\tilde\fn_i\|_{\rm F}$, where $\tilde\fn_i$ are generated by different Gaussian kernels with standard deviation $3$ and $5$, respectively}
\label{fig_SmoothNormals}
\end{figure*}

\begin{figure*}[t]
\centering
\includegraphics[width=17.5cm]{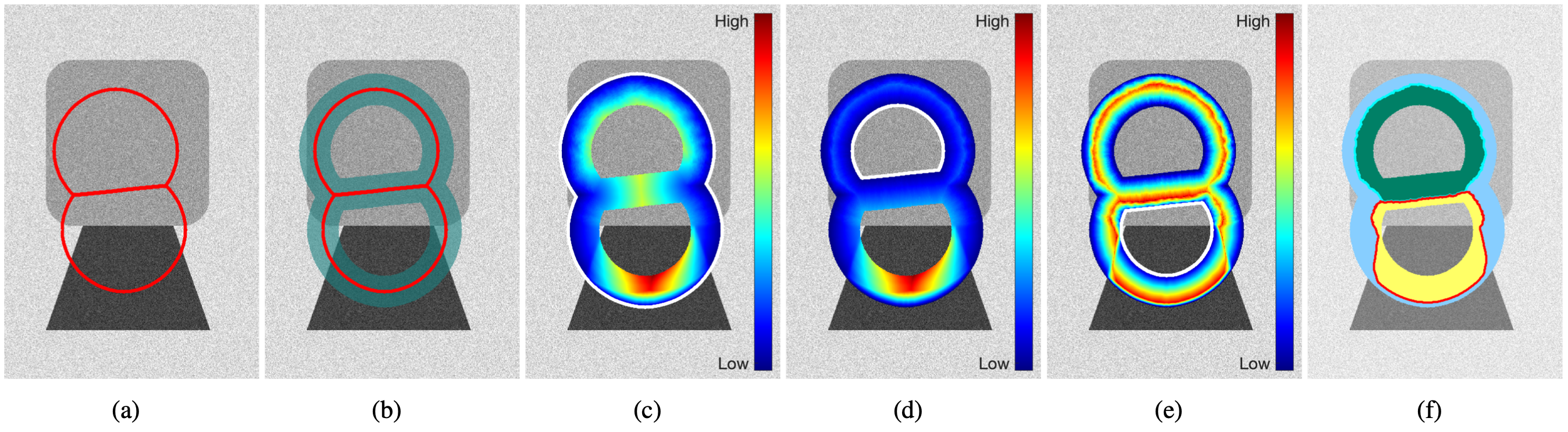}
\caption{An example for one step of the proposed Voronoi diagram-based dual-front model. (\textbf{a}) The original image with a given contour $\Gamma$ which is indicated by red lines. (\textbf{b}) The shadow region stands for its neighbourhood $U_\Gamma$. (\textbf{c}) to (\textbf{e}) Geodesic distance maps $\rU^{(1)}$, $\rU^{(2)}$ and $\rU^{(3)}$, respectively. The white lines are the offset lines $\cC_i^\ell$ for $i=1,\,2,\,3$.  (\textbf{f}) The corresponding Voronoi regions indicated by different colors}
\label{fig:Steps}
\end{figure*}

\section{Implementation Consideration}
\label{Sec:Numerical}
 
\subsection{Smooth the Vector Fields $\fn_i$}
\label{eq:SmoothVF}
The vector fields $\fn_i$ dominate the front propagation. In this section, we smooth $\fn_i$ by means of a Gaussian kernel in order to alleviate the effects from discretization. For this purpose, we first consider the following matrix field
\begin{equation}
\label{eq:AsyST}
\cT_i(x)=\left(G_a\ast \fn_i\fn_i^T\right)(x),
\end{equation}
where $G_a$ is a Gaussian kernel with standard deviation $a$.

Let us denote by $\varpi(x)$ the eigenvector of the matrix $\cT_i(x)$ which corresponds to the largest eigenvalue of $\cT_i(x)$. Then the smoothed vector fields, denoted by $\tilde\fn_i$, can be generated as follows
\begin{equation}
\label{eq:SmoothNormals}
\tilde\fn_i(x):=\langle\fn_i(x),\varpi(x)\rangle \varpi(x).
\end{equation}
From equation~\eqref{eq:SmoothNormals}, we can see that at each point $x\in\Omega$ the smoothed vector $\tilde\fn_i(x)$ actually forms an acute angle with the original one $\fn_i(x)$.

We apply the Frobenius norms of the Jacobian matrix of the vector field to illustrate its smoothness property~\cite{li2005segmentation}. 
Let us respectively denote by $\nabla\fn_i$ and  $\nabla\tilde\fn_i$ the Jacobian matrix fields of $\fn_i$ and $\tilde\fn_i$,  such that low values of the Frobenius norms $\|\nabla\fn_i(x)\|_{\rm F}$ (resp. $\nabla\tilde\fn_i(x)$) indicate slowly-varying vectors $\fn_i(x)$ (resp. $\tilde\fn_i(x)$).  We exploit a synthetic image to visualize the Frobenius norms $\|\nabla\fn_i\|_{\rm F}$ and $\|\nabla\tilde\fn_i\|_{\rm F}$, as depicted in Fig.~\ref{fig_SmoothNormals}. Fig.~\ref{fig_SmoothNormals}a illustrates the synthetic image with an initial contour $\Gamma$ (indicated by a red line), by which we can build its neighbourhood $U_\Gamma$ and establish the vector fields $\fn_i$ and $\tilde\fn_i$. The velocity function $\xi_{\rm ext}$ is  estimated using the piecewise constants model, where $\xi_{\rm ext}(x)<0$ if $x$ is outside $\Gamma$ and $\xi_{\rm ext}(x)>0$, otherwise.  In Fig.~\ref{fig_SmoothNormals}b, we visualize the Frobenius norms  $\|\nabla\fn_i(x)\|_{\rm F}$, where the red dash line denotes the initial contour $\Gamma$. The Frobenius norms $\|\nabla\tilde\fn_i(x)\|_{\rm F}$ shown in Figs.~\ref{fig_SmoothNormals}b and~\ref{fig_SmoothNormals}c are generated by two Gaussian kernels $G_a$ with standard deviations $a=3$ and $a=5$, respectively.  We can see that the values $\|\nabla\tilde\fn_i(x)\|_{\rm F}$ at points $x$ nearby the contour segment of weak tortuosity are lower than $\|\nabla\fn_i(x)\|_{\rm F}$, due to the use of the Gaussian smooth operation in Eq.~\eqref{eq:AsyST}. 

\begin{figure*}[t]
\centering
\subfigure[]{\includegraphics[height=4.3cm]{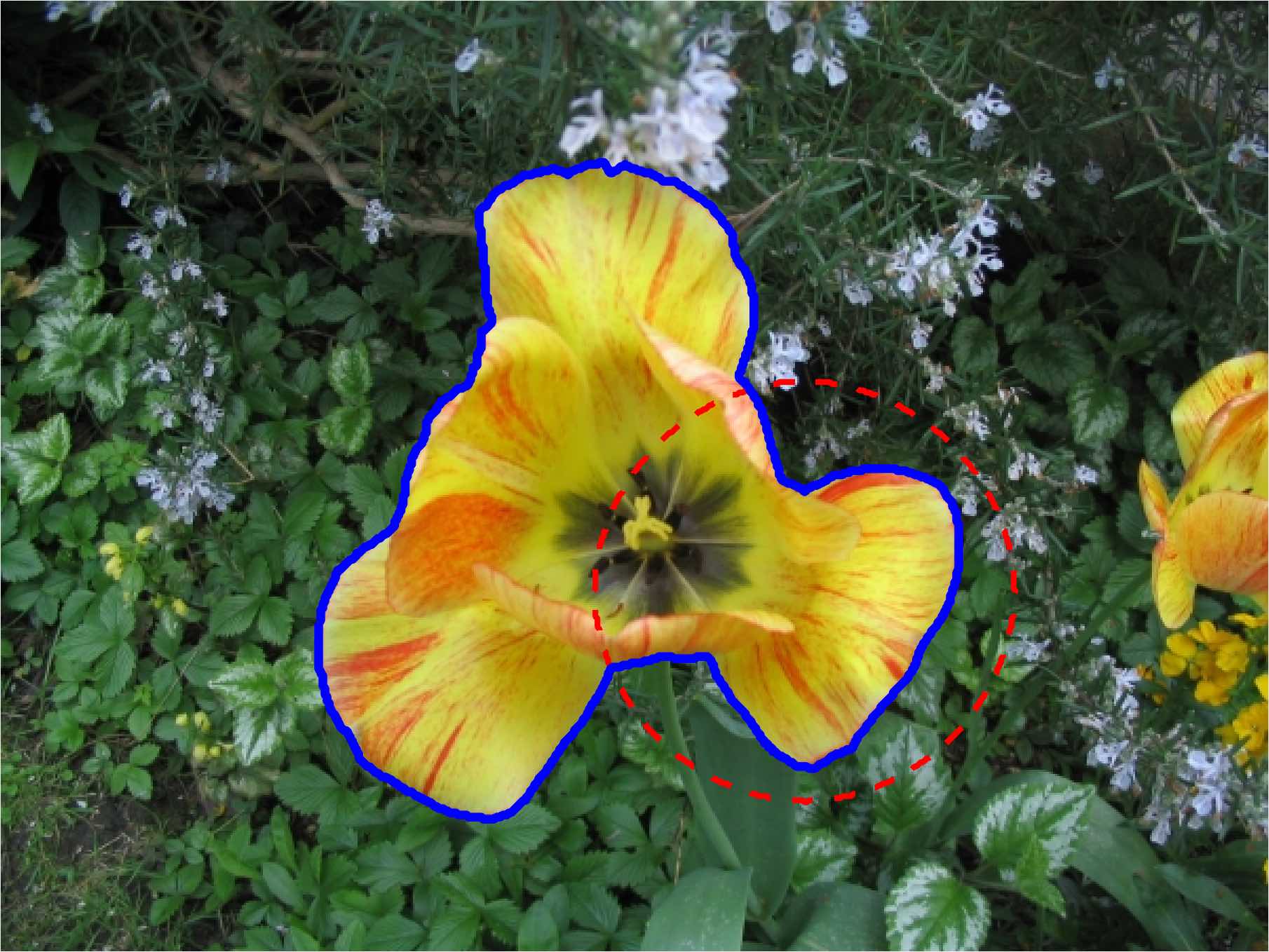}	}\,\subfigure[]{\includegraphics[height=4.3cm]{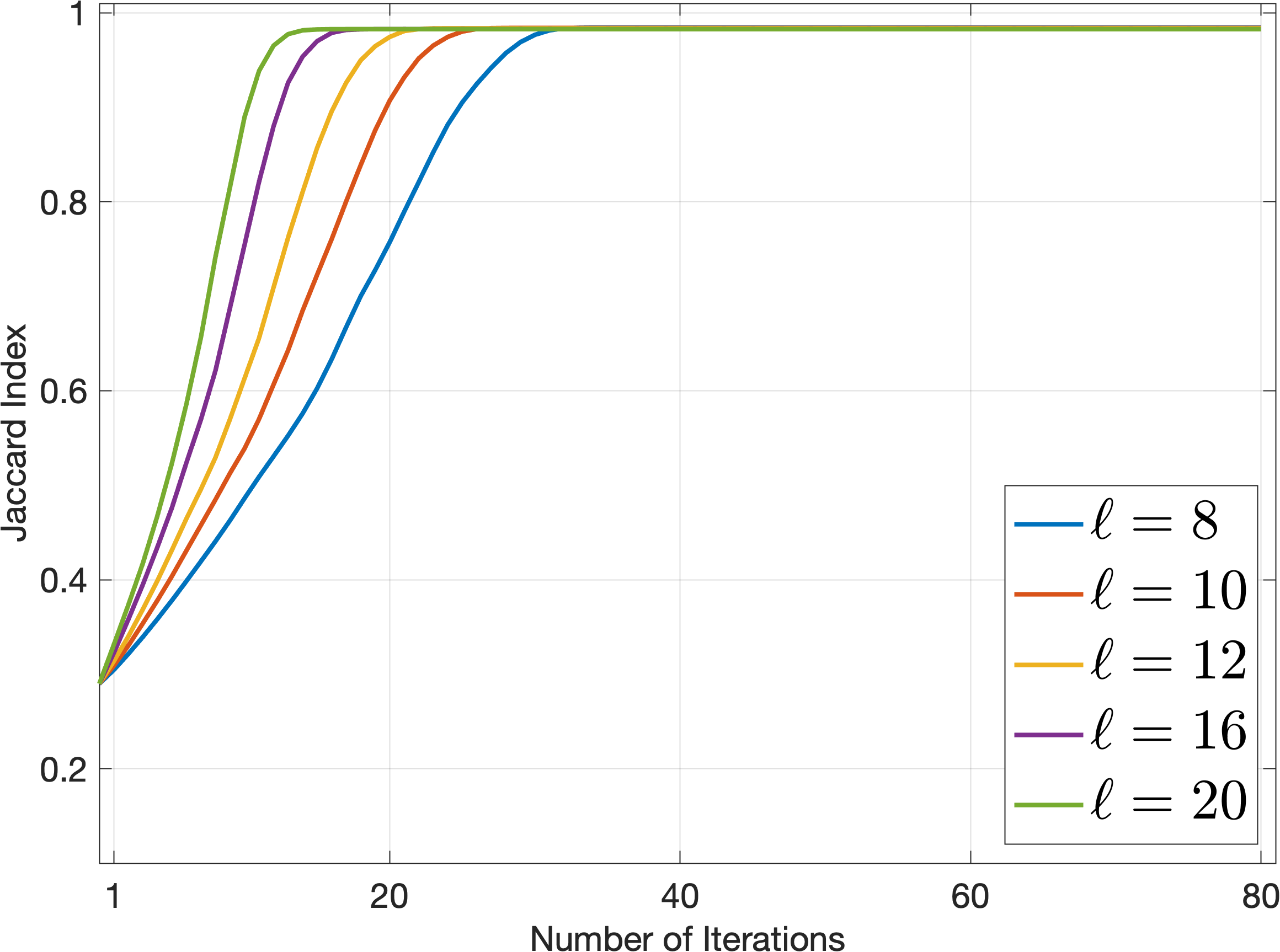}}\,\subfigure[]{\includegraphics[height=4.3cm]{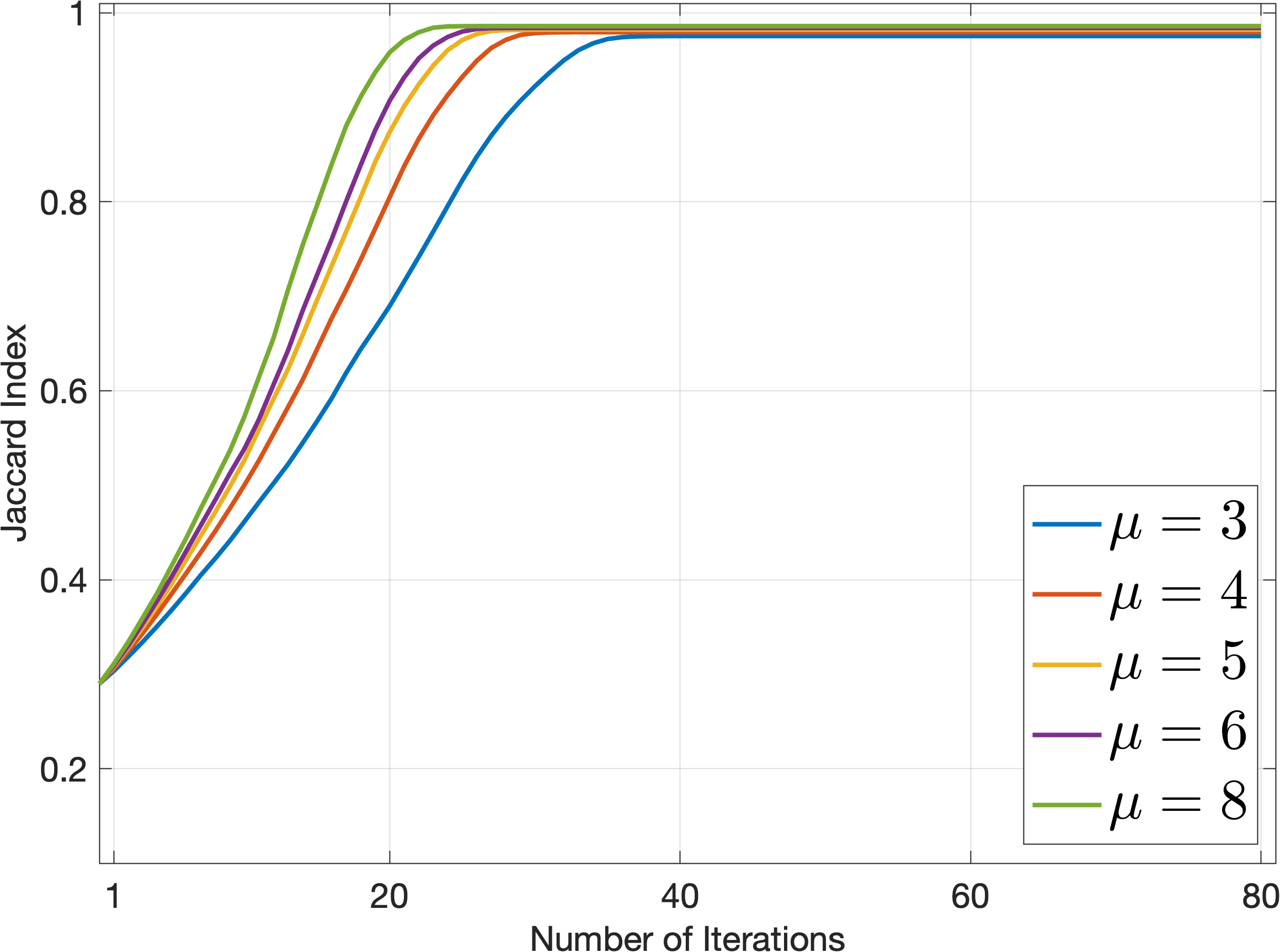}}
\caption{Convergence rate for different values of parameters. (\textbf{a}) Original image. The blue solid line is the ground truth contour and the red dash line indicates the initial contour. (\textbf{b}) and (\textbf{c}) Convergence rate corresponding to different values of $\ell$ and $\mu$, respectively}
\label{fig_Parameters}
\end{figure*}

\begin{algorithm}[!t]
\caption{Fast Marching Method with Prescribed Distances}	
\label{algo:SuccessiveFM}
\begin{algorithmic}
\renewcommand{\algorithmicrequire}{\textbf{Input:}}
\renewcommand{\algorithmicensure}{\textbf{Output:}}
\Require  A set $X_h\subset\Omega_h$ of source points, a metric $\kF$, an active region $U_h\subset\Omega_h$, and a given distance map $\rU$.
\Ensure Geodesic distance map $\rD$.
\end{algorithmic}
\begin{algorithmic}[1]
\State Set $\rD(x)=0,\,\forall x\in X_h$ and $\rD(x)=\infty,\,\forall x\in U_h\backslash X_h$.
\State Tag each grid point $x\in U_h$ as \emph{Trial}.
\State Build the stencils $\Lambda$ in terms of the metric $\kF$.
\While{there exists at least one point is tagged as \emph{Trial}}
\State Find a point $x_{\rm m}$ minimizing $\rD$ among all \emph{Trial} points.
\State Tag the point $x_{\rm m}$ as \emph{Accepted}.
\For{any \emph{Trial} point $y\in U_h$ s.t. $x_{\rm m}\in\Lambda(y)$}
\If{$\rU(y)\geq\rD(x_{\rm m})$ and $y\in U_h$} 
\label{alg2_Start}
\State Update $\rD(y)$ based on a subset of stencil $\Lambda(y)$.
\EndIf
\label{alg2_End}
\EndFor 
\EndWhile
\end{algorithmic}
\end{algorithm}

\subsection{Examples for Region-based Homogeneity Criteria}
We take region competition model~\cite{zhu1996region}, the Chan-Vese model~\cite{chan2000active, chan2001active}, and the Bhattacharyya coefficient model~\cite{michailovich2007image} as examples to derive the extended velocity functions (see Eqs.~\eqref{eq:CMotion} and \eqref{eq:ExtVelocity}), which are the crucial ingredients for the proposed dual-front model. 

In the region competition model~\cite{zhu1996region}, the image gray levels or colors in each region $\cR_i$ are supposed to follow a prescribed probability distribution such as the Gaussian distribution or more general the Gaussian mixture model. The Chan-Vese model~\cite{chan2001active} is a piecewise constants reduction of the full Mumford-Shah functional~\cite{mumford1989optimal}, which exploits a single Gaussian probability density function  to characterize the regional homogeneity measure in each $\cR_i$. In addition, both probability density functions are supposed to share an identical standard deviation value.
Moreover, the active contour model based on the Bhattacharyya coefficient~\cite{michailovich2007image} is a non-parametric segmentation approach. In the context of foreground and background segmentation, image segmentation is achieved by maximizing the discrepancy between the histograms of image features inside and outside the evolving contour. As a consequence, the priors on the image data distributions, as in the region competition model, are no longer required. In the following experiments,  we only consider the case of two-phase segmentation, i.e. the number of subregions is set to $n=2$, when computing the velocity functions through the Bhattacharyya coefficient-based functional. We make use of the Gaussian kernels for the construction of the histograms in each region and the band width for the Gaussian kernel is fixed to $2$. Finally, the extended velocity functions $\xi_i$ (for $1\leq i\leq n$) and $\xi_{\rm ext}$ associated to the models mentioned above are presented in Appendix~\ref{subsec:VelocityComput}.

\subsection{Hamiltonian Fast Marching for Distance Estimation}
Image segmentation based on the dual-front scheme is implemented through a contour evolution manner, as described in Algorithm~\ref{algo:Dual-Front}. In each evolution iteration, one of the key steps is to estimate a family of  geodesic distance maps in order to generate Voronoi regions in the neighbourhood of the input contour. We present the numerical implementation details for the estimation of distance maps with respect to the proposed asymmetric quadratic metrics, see Section~\ref{subsec_RegionalMetric}. 

In this paper, we make use of state-of-the-art Hamiltonian fast marching (HFM) method\footnote{The codes for the Hamiltonian fast marching  method can be downloaded from \href{https://github.com/Mirebeau/HamiltonFastMarching}{https://github.com/Mirebeau/HamiltonFastMarching}.}~\cite{mirebeau2018fast} as our numerical solver for the computation of geodesic distance maps. The HFM method is regarded as a generalization of the original fast marching method~\cite{sethian1999fast}. It can handle a wide variety of anisotropic and asymmetric Finsler metrics, in addition to classical isotropic Riemannian metrics. Numerically, the HFM method computes geodesic distances relying on a neighbourhood system $\Lambda$ generated in a regular grid $\Omega_h:=\Omega\cap\bZ^2$ with grid scale $h$. In our experiments, we set $h=1$. At each grid point $x\in\Omega_h$, the neighbourhood $\Lambda(x)$ is a finite set of grid points of $\Omega_h$, adaptively identified by a tensor decomposition procedure~\cite{mirebeau2019riemannian}. Such a set $\Lambda(x)$, also regarded as a stencil, collects all the neighbour grid points of $x$. In general, strongly anisotropic geodesic metrics may require stencils with large size in order to estimate accurate geodesic distances~\cite{mirebeau2014anisotropic,mirebeau2014efficient,mirebeau2019riemannian}. In the course of front propagation, the geodesic distance values are updated by solving the discretized Eikonal equation using an upwind finite difference scheme on a valid subset of the neighbourhood $\Lambda$. We refer to literature~\cite{mirebeau2019hamiltonian} for more details on the computation of geodesic distances.

In each iteration of contour evolution, given a contour $\Gamma$ as input, one can generate the neighbourhood $U_\Gamma$ of $\Gamma$ by Eq.~\eqref{eq:TubeNeigh}, $n$ narrow bands $U_i\subset U_\Gamma$ by Eq.~\eqref{eq:NeighBoundary}, and $n$ offset lines $\cC^\ell_i$ by Eq.~\eqref{eq:OffsetLines}. The output is a new contour made up of the interfaces between all adjacent Voronoi regions associated to the offset lines $\cC^\ell_i$. 
Each offset line $\cC^\ell_i$ is taken as the set of source points for the corresponding distance map $\rD_i$, i.e. $\rD_i(x)=0,\,\forall x\in \cC^\ell_i\cap\bZ^2$. All the maps $\rD_i$ for $1\leq i\leq n$ are estimated using a straightforward adaption of the HFM method in a successive manner~\cite{peyre2006geodesic}, such that the generation of all the Voronoi regions $\Vor(\cC_i^\ell)$ can be implemented in $n$ steps. For this purpose, we consider a prescribed distance map $\rU:U_\Gamma\cap\bZ^2\to\bR^+_0$, which serves as a constraint for the HFM. Specifically, when estimating each distance map $\rD_i$, only the distances at the grid points within a subset of $U_i$ need to be updated. This subset is related to $\rU$, as stated in Lines~\ref{alg2_Start} to~\ref{alg2_End} of Algorithm~\ref{algo:SuccessiveFM}. 

We denote by $\rU^{(i)}$ the updated map $\rU$ at the $i$-th step. At the initialization stage, we set $\rU^{(0)}(x)=\infty$ for each grid point $x\in U_\Gamma\cap\bZ^2$. Following that, in the $i$-th step ($i\geq 1$), the geodesic distance map $\rD_i$ is computed through Algorithm~\ref{algo:SuccessiveFM} by setting $\kF=\kF_i$, $X_h=\cC^\ell_i\cap\bZ^2$,  $U_h=U_i\cap\bZ^2$ and $\rU=\rU^{(i-1)}$ as inputs. Accordingly, the prescribed distance map $\rU^{(i)}$ can be updated as follows 
\begin{equation}
\label{eq:DistMap}
\forall x\in U_i\cap\bZ^2,\quad\rU^{(i)}(x)=\min\{\rU^{(i-1)}(x),\,\rD_i(x)\}.
\end{equation}

In order to construct the Voronoi regions $\Vor(\cC^\ell_i)$, we also estimate a Voronoi index map $\rV$, which assigns to each grid point $x\in\Vor(\cC_i^\ell)$ a label $i$. Similar to the update of the prescribed distance map $\rU$, the Voronoi index map at the $i$-th step, denoted by $\rV^{(i)}$, can be iteratively computed by
\begin{equation}
\label{eq_IndexMap}
\rV^{(i)}(x)=i, \quad\forall x\in U_i\cap \bZ^2\text{~s.t.~}\rD_i(x)<\Phi^{(i-1)}(x).
\end{equation}
Accordingly, one can build the Voronoi regions $\Vor(\cC^\ell_i)$ as follows
\begin{equation}
\label{eq:IndexMap}
\Vor(\cC^\ell_i):=\big\{x\in U_i\cap \bZ^2;\rV(x)=i\big\}.
\end{equation}

In Fig.~\ref{fig:Steps}, we illustrate an example for one iteration of the proposed dual-front model.  The test image from the GrabCut dataset~\cite{rother2004grabcut} is presented in Fig.~\ref{fig:Steps}a,  where the red line denotes the given contour $\Gamma$. The shadow region represents the neighbourhood $U_\Gamma$, as depicted in Fig.~\ref{fig:Steps}b. Figs.~\ref{fig:Steps}c to~\ref{fig:Steps}e respectively illustrate the geodesic distance maps $\rU^{(1)}$, $\rU^{(2)}$ and $\rU^{(3)}$ superimposed on the original image, with  white lines representing the offset lines $\cC^\ell_i$ for $i=1,2,3$. Note that in these figures, the distance maps $\rU^{(i)}$ are linearly normalized such that $\sup_x\{\rU^{(i)}(x)\}=1$ for better visualization.
In Fig.~\ref{fig:Steps}f, the constructed Voronoi regions $\Vor(\cC_1^\ell)$, $\Vor(\cC_2^\ell)$ and $\Vor(\cC_3^\ell)$ are illustrated by different colors. The cyan and red lines are the reconstructed interfaces between the corresponding adjacent Voronoi regions.

\section{Experimental Results}
\label{sec_Experiment}
In this section,  we illustrate the experimental results of the proposed dual-front model. The experiments  involve not only  the study of the properties of the proposed model itself, but also the qualitative and  quantitative comparisons with the classical dual-front model~\cite{li2007local}.

\subsection{Parameter Setting}
The parameter $\ell$ controls the thickness of the neighbourhood $U_\Gamma$. In each iteration of dual-front scheme as in Algorithm~\ref{algo:Dual-Front}, this neighbourhood serves as a searching space for the interface of Voronoi regions. In Fig.~\ref{fig_Parameters}b, we respectively examine the convergence rate associated to the width values $\ell=8,\,10,\,12,\,16$ and $20$ on an image from the GrabCut dataset~\cite{rother2004grabcut}. The initial contour overlapped on the original image is shown in Fig~\ref{fig_Parameters}a, where the region-based homogeneity term in this experiment is derived from the Bhattacharyya coefficient~\cite{michailovich2007image}. The convergence rate is evaluated in terms of the Jaccard index, or the Jaccard score $\cJ$, defined through the overlap part between the segmentation region $\Im$ and the ground truth $\GT$
\begin{equation}
\label{eq:Jaccards}
\cJ(\Im,\GT)=\frac{|\Im\cap \GT|}{|\Im\cup \GT|},	
\end{equation}
where $|\Im|$ stands for the area of the region $\Im$. From Fig.~\ref{fig_Parameters}b we can see that higher values of $\ell$ are capable of yielding contour convergence in less iterations. However,  the use of a high value of $\ell$ may give rise to unexpected segmentations. As a tradeoff, we set $\ell\in\{5,10\}$ in the following experiments, depending on the sizes of the tested images. 

The parameters $\mu$ and $\alpha$ control the relative importance of the region-based homogeneity penalty, where $\mu$ in Eq.~\eqref{eq:omega} dominates the asymmetric penalization of each geodesic metrics $\kF_{i}$.  We illustrate in Fig.~\ref{fig_Parameters}c the relationship between the convergence rate of the contour evolution and the values of the parameter $\mu$.  In Fig. \ref{fig_Parameters}c, we plot the values of Jaccard index values $\cJ$ for the proposed dual-front method with respect to different values of $\mu$. One can point out that high values of $\mu$ lead to fast convergence rate for the evolving contour. However, high values of $\mu$ will yield stencils of large size, which may reduce the locality of these stencils and increases the numerical cost of the HFM method.
The weighted functions $\psi_i$ are able to speed up the convergence of the evolving contour, which are partially controlled by the parameter $\alpha$. In the following experiments, we make use of the values of  $\mu\in\{5,6\}$ and $\alpha\in\{0.1,0.2\}$ for the proposed dual-front model, unless otherwise specified.

The computation of the image gradients is the first step for the estimation of edge-based features, where we use $\sigma=1$ for the Gaussian kernel $G_\sigma$, see Eq.~\eqref{eq:Jacobian}. Following that we set  $\beta\in\{0,1\}$ and $\rho=4$, unless otherwise specified, for computing the edge-based tensor field $\cM$, see Eq.~\eqref{eq:SmoothedEdgeTensor}. In the case of $\beta=1$, instead of using the tensor field $\cM$ itself, we exploit the smoothed version $\tilde\cM$ as defined in Eq.~\eqref{eq:SmoothedEdgeTensor} to build the asymmetric quadratic metrics $\kF_i$. Note that when the edge anisotropy features are unreliable, we adopt an isotropic reduction of the tensor field~\eqref{eq:Eigen} by setting $\beta=0$.

\subsection{Comparative Image Segmentation Results}
We compare the proposed asymmetric dual-front model to the Li-Yezzi  dual-front model~\cite{li2007local} and the geodesic distance thresholding model~\cite{malladi1998real}. For fair comparison, we extend the isotropic metrics used in the Li-Yezzi dual-front model~\cite{li2007local} to an anisotropic case:
\begin{equation}
\kR_i(x,\fu)=\psi_i(x)\sqrt{\langle\fu,\tilde\cM(x)\fu\rangle},\quad i=1,\cdots,n,
\end{equation}
where $\tilde\cM$ that carries the smoothed edge-based features is defined in Eq.~\eqref{eq:SmoothedEdgeTensor}. The metric $\kR_i$ is a symmetric reduction of the proposed metric $\kF_i$ by setting $\omega_i\equiv\mathbf{0}$. For the Li-Yezzi dual-front model, the weighting functions $\psi_i$, as in Eq.~\eqref{eq:PSI}, are estimated by using the values $\alpha\in\{1,2\}$. The values $\ell$ defining the neighbourhood width are precisely identical in both dual-front models.

The geodesic distance thresholding model~\cite{malladi1998real} aims to search for image segmentations via some level set of a geodesic distance map $\rD$. In its original setting, the geodesic distance map is estimated using an isotropic metric and a segmented region $\Im_T=\{x\in\Omega;\rD(x)\leq T\}$ is the interior region of the $T$-level set line of $\rD$. In our experiments, we choose the value of $T$ as follows: 
\begin{equation}
T^*=\underset{T\in[T_{1},T_{2}]}{\arg\max}\,\mathcal{J}(\Im_T,\GT),
\end{equation}
 where $T_{1}$ and $T_{2}$ are two positive constants defined being such that $|\Im_{T_1}|=90\%|\GT|$ and $|\Im_{T_2}|=110\%|\GT|$. We extend the isotropic distance thresholding model~\cite{malladi1998real} to the asymmetric case by invoking an asymmetric quadratic metric  as follows
\begin{equation}
\label{eq:AsyEdge}
\kF_{\rm th}(x,\fu)=\sqrt{\langle\fu,\tilde\cM_{\rm th}\fu\rangle+\langle\fu,g(x)\mathfrak{p}(x) \rangle^2_-}.
\end{equation}
where $\mathfrak{p}:\Omega\to\bR^2$ is a vector field associated to the normalized edge appearance map $\eta$ as defined in Eq.~\eqref{eq_EdgeAppearance}. Specifically, we consider $\mathfrak{p}(x)=(\nabla G_\sigma\ast\eta)(x)/((\nabla G_\sigma\ast\eta)(x)+\iota)$ with $\iota\in\bR^+$ being a sufficiently small constant. For a point $x$ close to an image edge, the vector $\mathfrak{p}(x)$ points to an edge point from $x$. The tensor field $\tilde\cM_{\rm th}$ have the same eigenvectors with the smoothed tensor field $\tilde\cM$. Denoted by $\tilde\vartheta_1(x)$ and $\tilde\vartheta_2(x)$ the eigenvectors of $\tilde\cM(x)$, the tensor field $\tilde\cM_{\rm th}$ can be written as follows
\begin{equation}
\tilde\cM_{\rm th}(x)=\sum_{k=1}^{2}\tau_k(x)\tilde\vartheta_k(x)\tilde\vartheta_k(x)^T,
\end{equation}
where $\tau_1$ and $\tau_2$ are two scalar-valued functions, which are defined as  $\tau_1=\max\{\exp(\rho\,\eta_{\rm th})-\epsilon,\epsilon_0\}\tau_2$, and $ \tau_2=\max\{\exp(\beta\,\eta_{\rm th})-\epsilon,\epsilon_0\}$. The parameters $\epsilon$ and $\epsilon_0$ are two positive constants,  which are set being such that $\tau_1(x)$ and $\tau_2(x)$ are sufficiently small at the homogeneous region where $\eta(x)\approx0$. The  function $\eta_{\rm th}$ is generated by thresholding $\eta$ using a scalar value $T_{\rm edge}$ such that $\eta_{\rm th}(x)=\eta(x)$ if $\eta(x)\geq T_{\rm edge}$, and $\eta_{\rm th}(x)=0$, otherwise. The weighted function $g$ used in Eq.~\eqref{eq:AsyEdge} is set as $g(x)=\tau_2(x)$.  Numerically, we fix the parameters $\epsilon=1$, $\epsilon_0=0.02$, $\beta=2$, $T_{\rm edge}\in\{0.15,0.2\}$ and $\rho=8$ for the metric $\kF_{\rm th}$. Eventually, for the geodesic distance thresholding model, we adopt $\sigma=2$ for the Gaussian kernel $G_\sigma$ to compute the image gradients~\eqref{eq:Jacobian}, unless other specified.

\begin{figure*}[t]
\centering
\includegraphics[width=16cm]{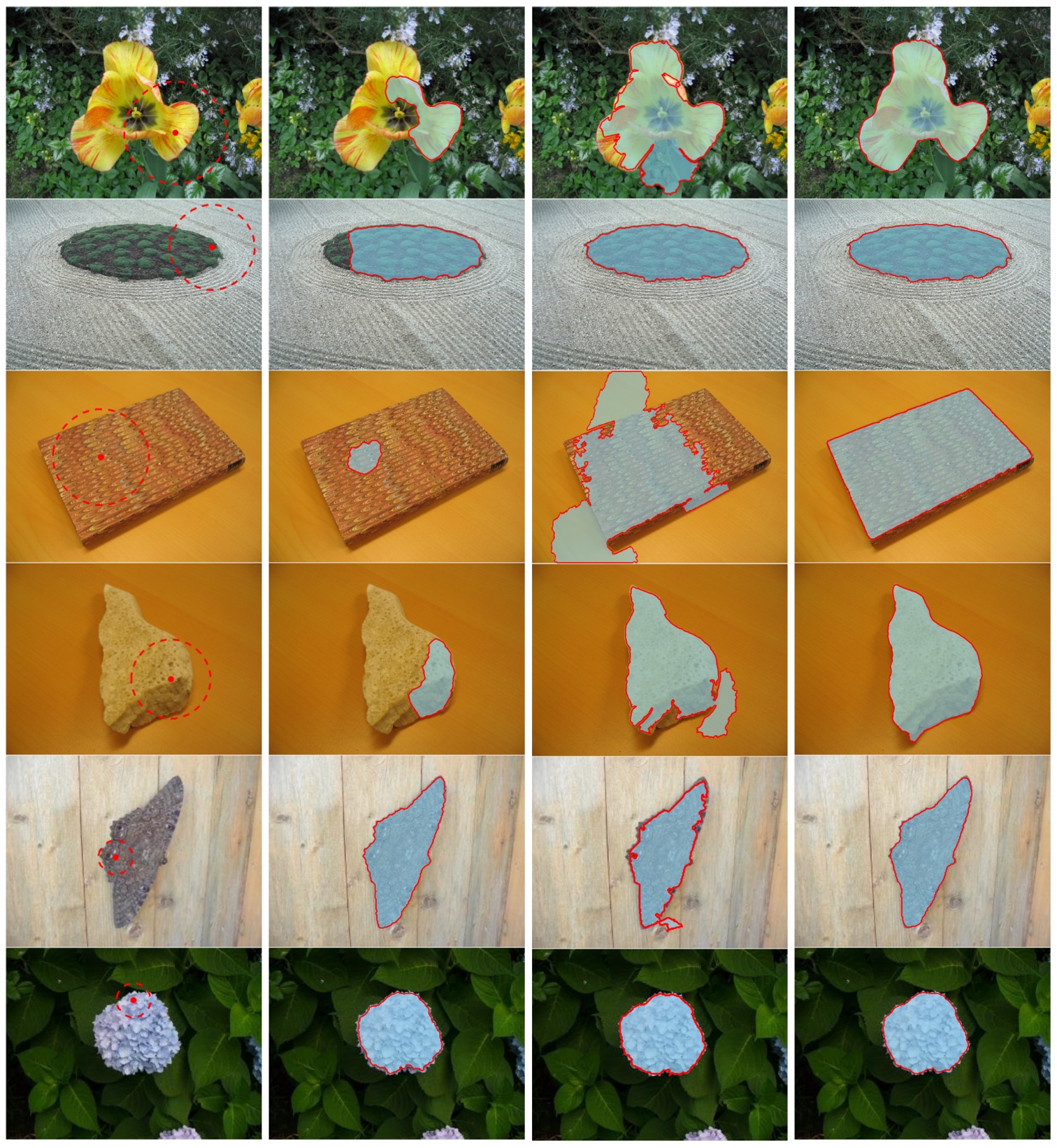}
\caption{Qualitative comparison with the Li-Yezzi dual-front model and the geodesic distance thresholding model. \textbf{Column} 1: Initial curves are indicated by red dash lines. \textbf{Columns} 2-4: Image segmentation results derived from the Li-Yezzi dual-front model, the geodesic distance thresholding model and the proposed dual-front model, respectively}
\label{Fig_NaturalImages}
\end{figure*}

In Fig. \ref{Fig_NaturalImages}, we illustrate the qualitative comparison results with  the Li-Yezzi dual-front model~\cite{li2007local} and the geodesic distance thresholding model on six tested images sampled from the Weizmann dataset and the Grabcut dataset.  In column $1$, the red dots are taken as the source points to perform the front propagation for the geodesic distance thresholding model. In this column, the initial curves for the Li-Yezzi dual-front model and the proposed model are  depicted by red dash lines. The image segmentation results derived from the Li-Yezzi dual-front model, the geodesic distance thresholding model and the proposed  model are demonstrated in columns $2$ to $4$, respectively. In the first $4$ rows of column $2$, one can see that the segmentation contours from the Li-Yezzi model pass through the interior regions of the target regions. In each of those rows, the interface of the Voronoi regions is stuck at unexpected positions. The segmentation results derived from the proposed  dual-front model are depicted in column $4$. One can point out that the use of the asymmetric quadratic metrics indeed yields segmentations capable of accurately depicting the  target boundaries. In column $3$, the segmentation results from the geodesic distance thresholding model are depicted, from which we one can observe that some portions of the final segmentation curves leak into the background, as shown in rows $1,3,4$ and $5$. Favorable segmentations are observed in rows $6$ for all the tested models, due to the well-defined image edges. For both  dual-front models, we exploit the Bhattacharyya coefficient model to compute the related region-based terms in rows $1$ to $4$, and the piecewise constant-model for the remaining  tests. The execution time (in seconds per evolution step) for the proposed dual-front model are  $0.54s$, $0.32s$, $0.64s$ and $0.61s$ with respect to the test images in rows $1$ to $4$. Note that in each evolution step, the execution times involve the estimation of the velocity functions, the construction of the neighbourhood regions, and the reconstruction of the interfaces of all adjacent Voronoi regions.  Reported execution times are obtained by running on a standard Intel Core i$9$ $3.6$GHz architecture with $96$Gb RAM.

\begin{figure}[t]
\centering
\includegraphics[height=9.5cm]{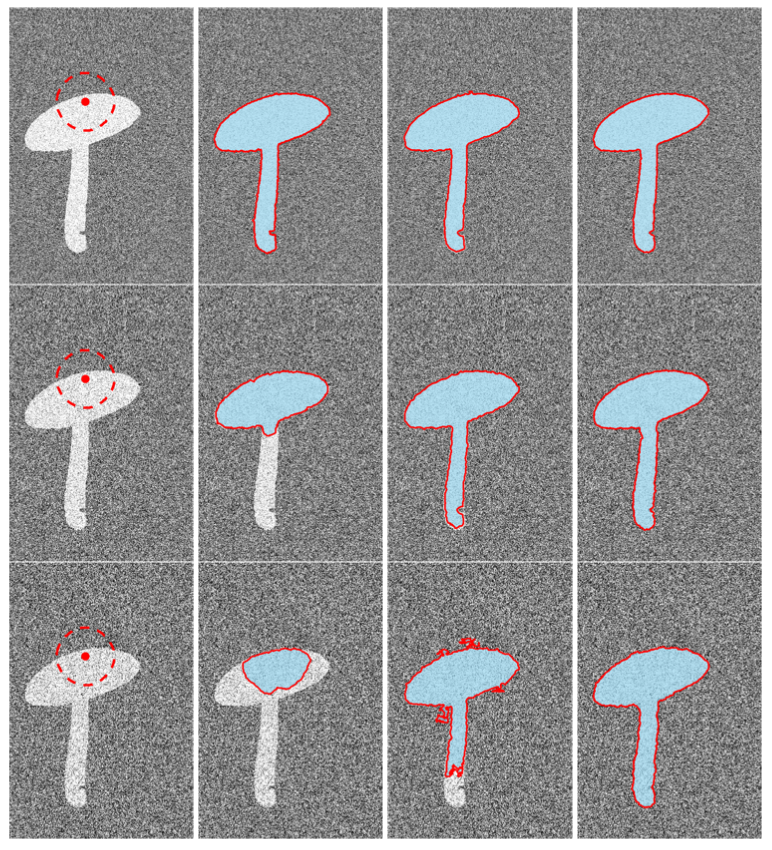}
\caption{Qualitative Comparison results on synthetic images blurred by different levels of noise. \textbf{Column} 1: The synthetic images with initial curves indicated bu red dash lines. \textbf{Columns} 2-4: Image segmentation results derived from the Li-Yezzi dual-front model, the geodesic distance thresholding model and the proposed asymmetric dual-front model, respectively}
\label{fig_syntheticImage}
\end{figure}

\begin{figure}[t]
\centering
\includegraphics[width=8cm]{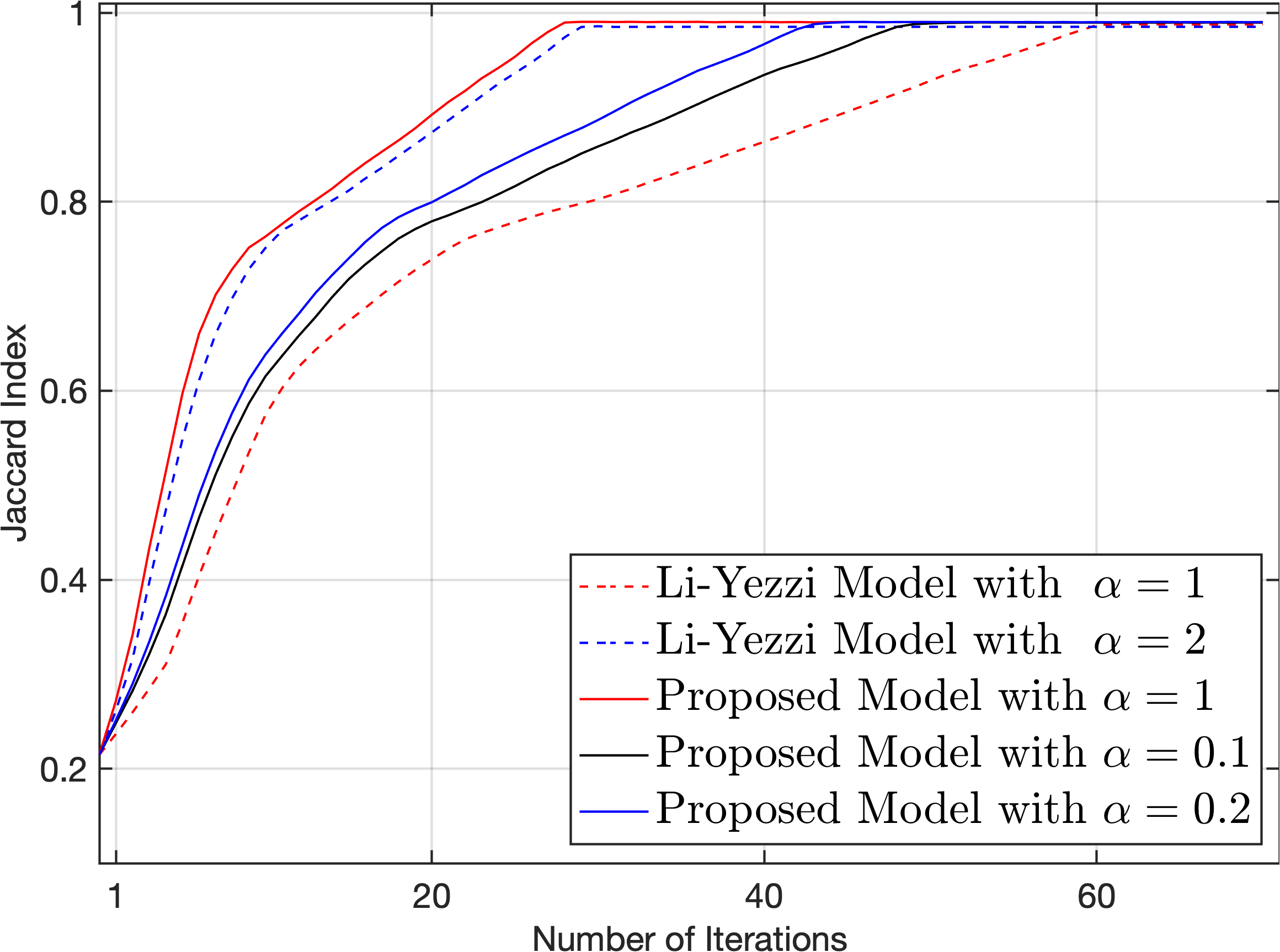}
\caption{Convergence rates for the Li-Yezzi dual-front model and the proposed model with respect to different values of the parameter $\alpha$. The tested image and the corresponding initial contour are shown in the top left of Fig.~\ref{fig_syntheticImage}}
\label{fig_ComConvergenceRate}	
\end{figure}

\begin{figure}[t]
\centering
\includegraphics[width=8cm]{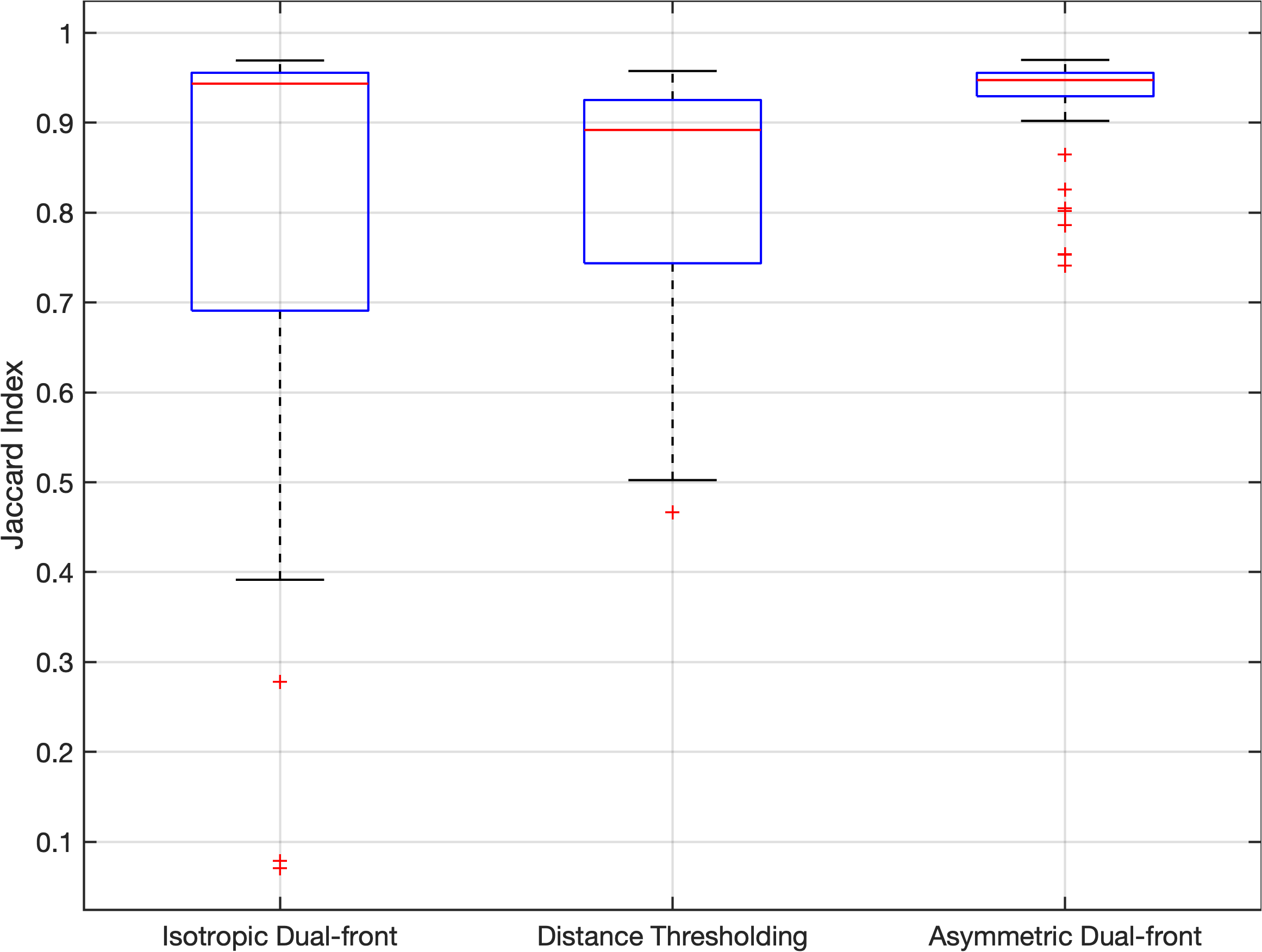}
\caption{Box plots of the Jaccard index values on $80$ CT image with respect to different models}
\label{fig_boxPlotCT}
\end{figure}

\begin{table*}[t]
\centering
\caption{Quantitative comparisons between the Li-Yezzi dual-front model, the geodesic distance thresholding model and the proposed asymmetric dual-front model in terms of the Jaccard index values (in percentage) evaluated over $20$ runs per image shown in Figs.~\ref{Fig_NaturalImages} and~\ref{fig_syntheticImage}}
\label{table:QuanResult}
\setlength{\tabcolsep}{5pt}
\renewcommand{\arraystretch}{1.5}
\begin{tabular}[t]{c c c c c c c c c c c c c c c}
\shline
\multicolumn{1}{c}{\multirow{1}{*}{Images}}&\multicolumn{4}{l}{Li-Yezzi Dual-front Model}&\multicolumn{4}{l}{Distance Thresholding Model}&\multicolumn{4}{l}{Asymmetric Dual-front Model}\\
\cmidrule(lr){2-5} \cmidrule(lr){6-9}\cmidrule(lr){10-13}
\multicolumn{1}{c}{}&\multicolumn{1}{c}{Ave}&\multicolumn{1}{c}{Max}&\multicolumn{1}{c}{Min}&\multicolumn{1}{l}{Std}&\multicolumn{1}{c}{Ave}&\multicolumn{1}{c}{Max}&\multicolumn{1}{c}{Min}&\multicolumn{1}{l}{Std}&\multicolumn{1}{c}{Ave}&\multicolumn{1}{c}{Max}&\multicolumn{1}{c}{Min}&\multicolumn{1}{l}{Std}\\
\hline
\multicolumn{1}{c}{Image 1}&\multicolumn{1}{c}{$25.5$}&\multicolumn{1}{c}{$48.4$}&\multicolumn{1}{c}{$16.4$}&\multicolumn{1}{l}{$0.06$}
&\multicolumn{1}{c}{$83.7$}&\multicolumn{1}{c}{$92.5$}&\multicolumn{1}{c}{$66.3$}&\multicolumn{1}{l}{$0.08$}
&\multicolumn{1}{c}{$90.1$}&\multicolumn{1}{c}{$98.1$}&\multicolumn{1}{c}{$20.0$}&\multicolumn{1}{l}{$0.24$}\\
\multicolumn{1}{c}{Image 2}&\multicolumn{1}{c}{$52.6$}&\multicolumn{1}{c}{$76.7$}&\multicolumn{1}{c}{$41.9$}&\multicolumn{1}{l}{$0.08$}
&\multicolumn{1}{c}{$95.1$}&\multicolumn{1}{c}{$95.7$}&\multicolumn{1}{c}{$91.7$}&\multicolumn{1}{l}{$0.01$}
&\multicolumn{1}{c}{$96.4$}&\multicolumn{1}{c}{$96.7$}&\multicolumn{1}{c}{$96.3$}&\multicolumn{1}{l}{$\approx0$}\\
\multicolumn{1}{c}{Image 3}&\multicolumn{1}{c}{$7.45$}&\multicolumn{1}{c}{$16.6$}&\multicolumn{1}{c}{$4.2$}&\multicolumn{1}{l}{$0.03$}
&\multicolumn{1}{c}{$23.2$}&\multicolumn{1}{c}{$50.4$}&\multicolumn{1}{c}{$6.7$}&\multicolumn{1}{l}{$0.12$}
&\multicolumn{1}{c}{$87.5$}&\multicolumn{1}{c}{$95.9$}&\multicolumn{1}{c}{$5.2$}&\multicolumn{1}{l}{$0.19$}\\
\multicolumn{1}{c}{Image 4}&\multicolumn{1}{c}{$12.0$}&\multicolumn{1}{c}{$21.9$}&\multicolumn{1}{c}{$8.0$}&\multicolumn{1}{l}{$0.04$}
&\multicolumn{1}{c}{$63.1$}&\multicolumn{1}{c}{$94.1$}&\multicolumn{1}{c}{$8.7$}&\multicolumn{1}{l}{$0.29$}
&\multicolumn{1}{c}{$71.1$}&\multicolumn{1}{c}{$97.2$}&\multicolumn{1}{c}{$21.0$}&\multicolumn{1}{l}{$0.23$}\\
\multicolumn{1}{c}{Image 5}&\multicolumn{1}{c}{$95.6$}&\multicolumn{1}{c}{$97.4$}&\multicolumn{1}{c}{$88.9$}&\multicolumn{1}{l}{$0.03$}
&\multicolumn{1}{c}{$92.5$}&\multicolumn{1}{c}{$93.8$}&\multicolumn{1}{c}{$90.4$}&\multicolumn{1}{l}{$0.01$}
&\multicolumn{1}{c}{$96.7$}&\multicolumn{1}{c}{$97.5$}&\multicolumn{1}{c}{$95.2$}&\multicolumn{1}{l}{$0.01$}\\
\multicolumn{1}{c}{Image 6}&\multicolumn{1}{c}{$91.2$}&\multicolumn{1}{c}{$94.9$}&\multicolumn{1}{c}{$27.4$}&\multicolumn{1}{l}{$0.15$}
&\multicolumn{1}{c}{$86.1$}&\multicolumn{1}{c}{$90.2$}&\multicolumn{1}{c}{$73.2$}&\multicolumn{1}{l}{$0.04$}
&\multicolumn{1}{c}{$94.1$}&\multicolumn{1}{c}{$97.2$}&\multicolumn{1}{c}{$61.7$}&\multicolumn{1}{l}{$0.08$}\\
\multicolumn{1}{c}{Synthetic 1}&\multicolumn{1}{c}{$91.3$}&\multicolumn{1}{c}{$98.5$}&\multicolumn{1}{c}{$28.4$}&\multicolumn{1}{l}{$0.22$}
&\multicolumn{1}{c}{$97.1$}&\multicolumn{1}{c}{$97.3$}&\multicolumn{1}{c}{$96.8$}&\multicolumn{1}{l}{$\approx 0$}
&\multicolumn{1}{c}{$98.7$}&\multicolumn{1}{c}{$99.1$}&\multicolumn{1}{c}{$96.8$}&\multicolumn{1}{l}{$\approx 0$}\\
\multicolumn{1}{c}{Synthetic 2}&\multicolumn{1}{c}{$56.2$}&\multicolumn{1}{c}{$91.4$}&\multicolumn{1}{c}{$16.2$}&\multicolumn{1}{l}{$0.26$}
&\multicolumn{1}{c}{$96.2$}&\multicolumn{1}{c}{$96.9$}&\multicolumn{1}{c}{$95.5$}&\multicolumn{1}{l}{$\approx 0$}
&\multicolumn{1}{c}{$97.4$}&\multicolumn{1}{c}{$97.8$}&\multicolumn{1}{c}{$97.2$}&\multicolumn{1}{l}{$\approx 0$}\\
\multicolumn{1}{c}{Synthetic 3}&\multicolumn{1}{c}{$18.2$}&\multicolumn{1}{c}{$31.5$}&\multicolumn{1}{c}{$5.4$}&\multicolumn{1}{l}{$0.06$}
&\multicolumn{1}{c}{$82.7$}&\multicolumn{1}{c}{$92.4$}&\multicolumn{1}{c}{$54.6$}&\multicolumn{1}{l}{$0.08$}
&\multicolumn{1}{c}{$96.9$}&\multicolumn{1}{c}{$97.0$}&\multicolumn{1}{c}{$96.0$}&\multicolumn{1}{l}{$\approx 0$}\\
\shline
\end{tabular}
\end{table*}

\begin{figure}[t]
\centering
\includegraphics[width=9cm]{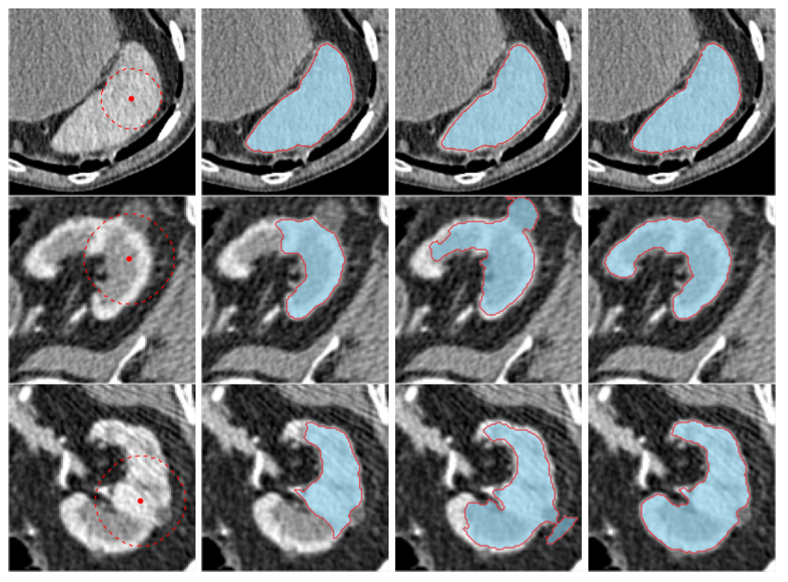}
\caption{Sampled examples from the dataset of CT images, where the quantitative comparison results are illustrated in Fig.~\ref{fig_boxPlotCT}. \textbf{Column} 1: The original CT images with initial curves (dash lines). \textbf{Columns}  2-4: Image segmentation results from the Li-Yezzi model, the geodesic distance thresholding model and the proposed model, respectively}
\label{fig_CTExamples}
\end{figure}

\begin{figure*}[t]
\centering
\includegraphics[width=17.5cm]{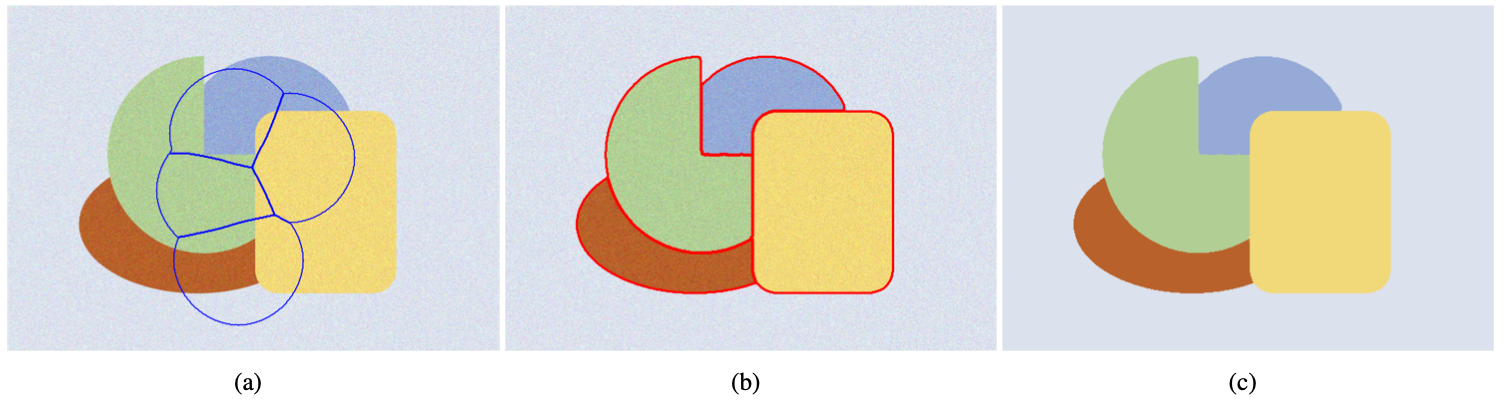}
\caption{An example for multi-region segmentation by the proposed asymmetric dual-front model. (\textbf{a}) A synthetic image with initial curves (blue lines) which partition the image domain to $5$ subregions. (\textbf{b}) The segmentation contour represented by red lines. (\textbf{c}) Mean color values in each segmented region}
\label{Fig_MultiRegion}
\end{figure*}

In Fig.~\ref{fig_syntheticImage}, we evaluate the performance of the three tested models mentioned above on synthetic images interrupted by different levels of adaptive Gaussian noise. The initial curves for each  synthetic image are shown in column $1$, where the  noise levels increase from rows $1$ to $3$. The segmentation results in columns $2$ to $4$ are respectively generated from the Li-Yezzi dual-front model, the geodesic thresholding model and the proposed model.  We can see that the image segmentation contour depicted in row $2$ and column $2$, derived from the Li-Yezzi model, misses the elongated part of the target region. While in row $3$ and column $2$, the segmentation contour suffers from a serious shortcut problem. In the first two rows of column $3$, the segmentation contours are generated using the geodesic distance thresholding model, which are able to well capture the target regions. However, the geodesic distance thresholding model relying on the image gradient-based features are  sensitive to the influence of image noise, as described in rows $2$ and $3$. The segmentations shown in column $4$ of Fig.~\ref{fig_syntheticImage} are obtained from the proposed asymmetric dual-front model. One can observe favorable segmentation results thanks to the integration of the image features and the asymmetry enhancement. In this experiment, we use the Gaussian mixture model to  derive the region-based homogeneity features for both dual-front models.

In Table.~\ref{table:QuanResult}, we show the quantitive comparisons between  the Li-Yezzi dual-front model, the geodesic distance thresholding model and the proposed asymmetric dual-front model on the images shown in Figs.~\ref{Fig_NaturalImages} and~\ref{fig_syntheticImage}. Those models are performed for $20$ times on each tested image. We first construct a set $\Re=:\{x_1,\cdots,x_{20}\}$ comprised of  $20$ grid points inside the eroded ground truth, obtained by a farthest point sampling scheme~\cite{peyre2010geodesic}, see Appendix~\ref{appendix:FPS}. For both dual-front models, the initial contour in the $k$-th test are set as a circle centered as the $k$-th grid point $x_k\in\Re$. For the geodesic distance thresholding model, we directly use the point $x_k\in\Re$ as the source point for the $k$-th test. We compute the statistics involving the average (Avg.), maximum (Max.), minimum (Min.) and standard derivation (Std.) values of the Jaccard index $\cJ$ with respect to the conducted $20$ tests.  It appears that the Ave. Jaccard scores for Li-Yezzi model for images $1$ to $4$ as well as synthetic images $2$ to $3$ exhibit poor segmentations, as can be seem from Table.~\ref{table:QuanResult}. By simultaneously taking into account the Ave. scores and the results shown in Figs.~\ref{Fig_NaturalImages} and~\ref{fig_syntheticImage}, the final contour derived from the Li-Yezzi model may only capture a small part of each target boundary. The statistics of Jaccard scores for the geodesic thresholding model are in general better than those from the Li-Yezzi model. However, we still observe that the Ave. scores for the geodesic thresholding model are less than $85\%$ in some test images, mainly because of the significant leaking problem. One can see that the asymmetric dual-front model indeed achieves the highest Avg. and Max. scores than the other compared models, proving the effectiveness of the asymmetric penalty in the proposed asymmetric dual-front model.

Among the experiments conducted above, we have respectively chosen the parameter $\alpha\in\{1,2\}$ to set up the Li-Yezzi dual-front model, and lower values of $\alpha\in\{0.1,0.2\}$ for the proposed asymmetric dual-front model, in order to demonstrate the advantages  of using the introduced asymmetric quadratic metrics. In Fig.~\ref{fig_ComConvergenceRate}, we illustrate the comparison results on the convergence rate of both dual-front models, using the test image shown in the first row of Fig.~\ref{fig_syntheticImage}. In this experiment, the convergence rates are characterized by the varying Jaccard index values $\cJ$ with respect to the number of contour iterations.  Even through given lower values of the parameter $\alpha$, one can point out that the convergence rates corresponding to the proposed model (indicated by solid lines) are indeed faster than the Li-Yezzi model (indicated by dash lines), due to the existence of asymmetry penalization encoded in the asymmetric quadratic metrics considered.

Eventually, we evaluate the Li-Yezzi dual-front model, the geodesic distance thresholding model and the proposed asymmetric dual-front model on $80$ CT images~\cite{spencer2019parameter}. In this experiment, the initial contour in each image is a circle centred at an interior point that is farthest  to the boundary of  the ground truth region in the sense of Euclidean distance. The average values of $\cJ$ for the Li-Yezzi model, the geodesic distance thresholding model and the proposed model are respectively $81.0\%$, $83.6\%$ and $92.8\%$. Moreover, we exhibit the box plots in Fig.~\ref{fig_boxPlotCT} of the statistics of the Jaccard index values from those models. One can claim that the proposed asymmetric dual-front model indeed achieves the highest accuracy among all the compared models. In this experiment, we use the piecewise constant model to set up both dual-front models. We choose $\alpha=1$  for both dual-front models and $\mu=3$ for the proposed model. Furthermore, a small neighbourhood width $\ell=5$ is applied in this experiment for both dual-front models, due to the low resolution of the tested CT images. In Fig.~\ref{fig_CTExamples}, we illustrate the image segmentation results, which are produced by the evaluated models on three typical examples sampled from the CT dataset.

\noindent\emph{Discussion}. In the basic formulation of the dual-front scheme, the evolving contour is represented by the interfaces of all adjacent Voronoi regions. The foreground and background segmentation is a fundamental problem, for which the dual-front model can find suitable solutions. Moreover, the multi-region segmentation task can also be efficiently addressed by the dual-front model, as discussed in~\cite{dubrovina2015multi}. We show such an example in Fig.~\ref{Fig_MultiRegion} on a synthetic image, where in this case the initial contour $\Gamma$ is regarded as the union of a series of closed curves, which are depicted by blue lines in Fig.~\ref{Fig_MultiRegion}a. Fig.~\ref{Fig_MultiRegion}b shows the the final segmentation contour generated in several contour evolution steps and Fig.~\ref{Fig_MultiRegion}c illustrates the mean color values in each subregion. 
Finally, the proposed dual-front model can also be investigated to interactive  segmentation on 2D images and 3D volumes based on several user-provided scribbles. These scribbles, each of which can be regarded as a set of points, can provide reliable samples of image features, in order to estimate image data statistics within each subregion. We leave such an interactive segmentation adaption of the proposed dual-front model to the future work.

\section{Conclusion}
\label{sec_Conclusion}
In this paper, we introduce a new dual-front active contour model to address the image segmentation problems. The main contribution of this paper lies at the introduction of asymmetric quadratic metrics to the Voronoi diagram-based dual-front model. As a consequence, the proposed dual-front model is able to blend the benefits from both an asymmetry enhancement and the image region- and edge-based features. 
The asymmetric features of the considered metrics are derived from the predicted directions, which characterize the motion of the neighbouring offset lines associated to the evolving contour. Contrary to the classical Li-Yezzi model using Riemannian metrics, the introduced dual-front model with asymmetry enhancement is capable of alleviating the shortcut problem, thus can generate more accurate and robust segmentation results in various segmentation scenarios.  

\section*{Acknowledgment}
The authors would like to thank all the anonymous reviewers for their invaluable  suggestions to improve this manuscript. This work is in part supported by the National Natural Science Foundation of China (NOs. 61902224, 61906108), by the French government under management of Agence Nationale de la Recherche as part of the ``Investissements d'avenir'' program, reference ANR-19-P3IA-0001 (PRAIRIE 3IA Institute) and by new AI project towards the integration of education and industry in QLUT (NO. 2020KJC-JC01). The second author’s work was supported by the Wellcome Trust Institutional Strategic Support Award (204909/Z/16/Z). This research is also partially supported by the Young Taishan Scholars (NO.tsqn201909137).

\appendix
\label{sec:appendix}
\subsection{Convexity for the Asymmetric Quadratic Metric}
\label{subsec_Convexity}

We show that the metric $\cF(x,\fu)$ with a form of~\eqref{eq:AQMetric} is convex with respect to its second argument $\fu$. 
\begin{proposition}
\label{prop:Convexity}	
Let $M\in\bS^+_d$ be a positive definite symmetric matrix and let $\fw\in\bR^d$ be a vector, where $d=2,\,3$. Then the following function is convex on $\bR^d$. 
\begin{equation}
F(\fu):=\sqrt{\langle\fu,M\fu\rangle+\langle\fu,\fw\rangle_-^2},
\end{equation}
\end{proposition}

\begin{IEEEproof}
\renewcommand{\IEEEQED}{\IEEEQEDopen}
We denote by $F(\fu)=f(g_1(\fu),g_2(\fu))$, where $f$, $g_1$ and $g_2$ are the functions respectively defined as follows:
\begin{align*}
&f(a,b):=\sqrt{a^2+b^2},	\\
&g_1(\fu):=\sqrt{\langle\fu,M\fu \rangle},\quad g_2(\fu):=\max\{0,-\langle\fu,\fw \rangle\}.
\end{align*}
Clearly the functions $f$, $g_1$ and $g_2$ are convex. Specifically, $f$ is non-decreasing, componentwise, on the non-negative quadrant $(a,b)\in[0,\infty)$. Moreover, the functions $g_1$ and $g_2$ take non-negative values. The result follows, recalling that the composition of a convex non-decreasing function, with convex functions, defines a convex function, which concludes the proof.
\end{IEEEproof}

\subsection{Computation for Velocity Functions} 
\label{subsec:VelocityComput}
Let $\Gamma$ be a set of simple closed curves which partition the image domain $\Omega$ to $n$ subregions $\cR_i$. Here we denote by $\fI=(I_1,\cdots,I_M):\Omega\to\bR^M$ a gray level image for $M=1$ or a color image for $M=3$. 

\noindent\emph{Velocity functions from the region competition model}. In the region competition model, the region-based homogeneity property can be described via Gaussian mixture models. In this case, the PDF for the Gaussian mixture model in each region $\cR_i$ can be expressed as
\begin{equation}
\label{eq:GMM}
P_i(z;\Theta_i)=\sum_{k=1}^K \lambda_k \rG(z;\Theta_{i,k}),~ \Theta_i=(\Theta_{i,1},\cdots,\Theta_{i,K}),
\end{equation}
where $\forall\lambda_k\geq 0$ and $\sum_{k=1}^K\lambda_k=1$ are the weights for the $k$-th Gaussian distribution $\rG(z;\Theta_{i,k})$ with parameters $\Theta_{i,k}:=(\bm{c}_{i,k},\bm{\sigma}_{i,k})$. Specifically, $\bm{c}_{i,k}$ represents the mean values of image data  within the region $\cR_i$, and $\bm{\sigma}_{i,k}$ is the covariance matrix. These parameters $\Theta_i$ can be updated by using the Expectation Maximization algorithm.

With these definitions, the energy functional~\eqref{eq:Energy} for the region competition model can be reformulated as
\begin{equation*}
E(\Gamma)=\sum_{i=1}^n\int_{\cR_i}-\log\left(P_i(\mathbf{I}(x);\Theta_i)\right)dx.
\end{equation*}
Then the velocity functions $\xi_i$ for $1\leq i\leq n$ read as
\begin{equation*}	
\xi_i(x)=-\log\left((P_i(\mathbf{I}(x);\Theta_i)\right)
\end{equation*}
yielding that for any point $x\in\cR_i\cap \Vor(\Gamma_{i,j})$
\begin{equation*}	
\xi_{\rm ext}(x)=\xi_j(x)-\xi_i(x)=\log\left(\frac{P_i(\mathbf{I}(x);\Theta_i)}{P_j(\mathbf{I}(x);\Theta_j)}\right).
\end{equation*}

For the piecewise constant models~\cite{chan2000active,chan2001active}, the energy functional $E(\Gamma)$ can be simplified as  
\begin{equation*}
E(\Gamma)=\sum_{i=1}^n\int_{\cR_i}\|\fI(x)-\bm{c}_i\|^2 dx,	~\bm{c}_i=\left(c^{(1)}_i,\cdots,c^{(M)}_i\right),
\end{equation*}
where $c_i^{(m)}$ represents the mean intensity value of $I_m$ within the region $\cR_i$. In this case, the velocity functions $\xi_i$
\begin{equation*}
\xi_i(x)=\|\fI(x)-\bm{c}_i\|^2.
\end{equation*}
Then the velocity function $\xi_{\rm ext}$ can be estimated by Eq.~\eqref{eq:ExtVelocity}.

\noindent\emph{Velocity functions from Bhattacharyya coefficient}.
 The Bhattacharyya coefficient-based active contour model~\cite{michailovich2007image} has proven its strong ability in image segmentation. Basically, this model made use of the Bhattacharyya coefficient between pairs of  PDFs or histograms to construct the objective energy functional.
In the two-phase segmentation, the contour $\Gamma$ partitions the image domain $\Omega$ into two non-overlapped subregions $\cR_1$ and $\cR_2$, where we suppose $\cR_1$ is the interior region of $\Gamma$. The histogram of image features within each region $\cR_i$ often relies on a Gaussian kernel $G_i$, which can be written as
\begin{equation}
P_i(\pi;\Gamma)=\frac{1}{|\cR_i|}\int_{\cR_i}G_i(\pi-\fI(x))dx,
\end{equation}
where $|\cR_i|$ denotes the area of $\cR_i$.

In this case, the Bhattacharyya coefficient can be defined as
\begin{equation}
\label{eq:BhaCoeff}
\mathfrak{B}(\Gamma)=\int_\Pi\sqrt{P_1(\pi;\Gamma)P_2(\pi;\Gamma)}\,d\pi,
\end{equation}
where $\Pi$ denotes the feature space. 

The velocity functions $\xi_1$ and $\xi_2$ can be formulated as 
\begin{equation}
\xi_1(x)=-\frac{1}{2}\mathfrak{B}(\Gamma)(|\cR_1|^{-1}-|\cR_2|^{-1})+\frac{1}{2}\mathcal{Y}(x),
\end{equation}
and $\xi_2(x)=-\xi_1(x)$, where the term $\mathcal Y$ is defined as
\begin{align*}
\mathcal{Y}(x)=&\int_\Pi G_1(\pi-\fI(x))   
\left(\frac{1}{|\cR_1|}\sqrt{\frac{P_2(x;\Gamma)}{P_1(x;\Gamma)}}\right)d\pi\nonumber\\
&-\int_\Pi G_2(\pi-\fI(x))
\left(\frac{1}{|\cR_2|}\sqrt{\frac{P_1(x;\Gamma)}{P_2(x;\Gamma)}}\right)d\pi.
\end{align*}
In practice, one can set the kernels $G_1=G_2$ to simplify the computation, as discussed in~\cite{michailovich2007image}.

\subsection{Farthest Point Sampling}
\label{appendix:FPS}
We use the Euclidean distance-based farthest point sampling scheme~\cite{peyre2006geodesic} to get a set $\Re$ involving grid points  within a given connected  region $\GT\subset\bZ^2$. These points are expected to  distribute evenly in  $\GT$  as much as possible. 
For this purpose, we first randomly choose a grid point $x_1\in\GT$ and initialize the target $\Re=\{x_1\}$. Then we extract a point $x_2\in \GT$ that is farthest to $\Re$ in the sense of Euclidean distance
\begin{equation}
x_2=\underset{x\in\GT}{\arg\max}\left\{\min_{y\in\Re}\|x-y\|\right\}.
\end{equation}
Once $x_2$ is detected, we update the target set as $\Re=\{x_1,x_2\}$. One can repeat such a farthest point sampling procedure till the grid point $x_{N}$ is added to the set $\Re$.

\ifCLASSOPTIONcaptionsoff
 \newpage
\fi

\bibliographystyle{IEEEbib}
\bibliography{frontsPropagation}

\end{document}